\documentclass{article}
\usepackage{spconf}

\usepackage{multirow}
\usepackage{graphicx}
\usepackage{amsmath}
\usepackage[psamsfonts]{amssymb}
\usepackage{amsxtra}
\usepackage{threeparttable}
\usepackage{amsfonts}
\usepackage{mathrsfs}
\usepackage[caption=true, font=footnotesize,
nearskip=0.0pt,farskip=0.0pt, captionskip=0.0pt]{subfig}
\usepackage{arydshln}
\usepackage{cite}
\usepackage{enumitem}

\makeatletter
\renewcommand\section{\@startsection{section}{1}{\z@}
                      {0.5ex \@plus 0ex \@minus -2ex}
                      {0.5ex \@plus 0ex}
                      {\normalfont\Large\bfseries}}
\renewcommand\subsection{\@startsection{subsection}{2}{\z@}
                      {0.5ex \@plus 0ex \@minus -2ex}
                      {0.5ex \@plus 0ex}
                      {\normalfont\large\bfseries}}
\renewcommand\subsubsection{\@startsection{subsubsection}{3}{\z@}
                      {0.5ex \@plus 0ex \@minus -2ex}
                      {0.5ex \@plus 0ex}
                      {\normalfont\normalsize\bfseries}}
\def\@listi{\leftmargin\leftmargini
            \parsep 1.0pt
            \topsep 0.2\baselineskip \@minus 0.1\baselineskip
            \itemsep 1.0pt \relax}
\let\@listI\@listi
\makeatother

\title{Multi-Exposure Image Fusion Based on Exposure Compensation}
\name{Yuma Kinoshita$^{\star}$ \qquad Taichi Yoshida$^{\dagger}$ \qquad Sayaka Shiota$^{\star}$ \qquad Hitoshi Kiya$^{\star}$}
\address{$^{\star}$ Tokyo Metropolitan University, Tokyo, Japan \\
    $^{\dagger}$ Nagaoka University of Technology, Niigata, Japan}
\begin{document}\sloppy
\setlength{\parskip}{0.0pt}
\setlength{\tabcolsep}{1.0pt}
\setlength{\textfloatsep}{1.0pt}
\setlength{\floatsep}{0.0pt}
\setlength{\abovecaptionskip}{0.0pt}
\setlength{\belowcaptionskip}{1.0pt}
\setlength{\intextsep}{0.0pt}
\setlength{\dblfloatsep}{0.0pt}
\setlength{\dbltextfloatsep}{2.0pt}
\abovedisplayskip=0pt
\belowdisplayskip=0pt
\ninept
\maketitle
\begin{abstract}
  This paper proposes a novel multi-exposure image fusion
  method based on exposure compensation.
  Multi-exposure image fusion is a method to produce images without color saturation
  regions, by using photos with different exposures.
  However, in conventional works,
  it is unclear how to determine appropriate exposure values, and moreover, it is difficult
  to set appropriate exposure values at the time of photographing
  due to time constraints.
  In the proposed method, the luminance of the input multi-exposure images
  is adjusted on the basis of the relationship
  between exposure values and pixel values,
  where the relationship is obtained by assuming that a digital camera
  has a linear response function.
  The use of a local contrast enhancement method is also
  considered to improve input multi-exposure images.
  The compensated images are finally combined by one of
  existing multi-exposure image fusion methods.
  In some experiments, the effectiveness of the proposed method
  are evaluated in terms of the tone mapped image quality index,
  statistical naturalness, and discrete entropy,
  by comparing the proposed one with conventional ones.
\end{abstract}
%
%\renewcommand{\thefootnote}{\fnsymbol{footnote}}
%\footnote[0]{This research was (partly) supported by Grant-in-Aid for Research on
%  Priority Areas, Tokyo Metropolitan University, figsesearch on social big data."}
%\renewcommand{\thefootnote}{\arabic{footnote}}
%
\section{Introduction}
  The low dynamic range (LDR) of the imaging sensors used in modern digital cameras
  is a major factor preventing cameras from capturing images as good as those with human vision.
  Various methods for improving the quality of a single LDR image by enhancing the contrast
  have been proposed\cite{zuiderveld1994contrast, wu2017contrast, kinoshita2017pseudo}.
  However, contrast enhancement cannot restore saturated pixel values in LDR images.
  
  Because of such a situation, the interest of multi-exposure image fusion has
  recently been increasing.
  Various research works on multi-exposure image fusion have so far been reported
  \cite{goshtasby2005fusion,mertens2009exposure,saleem2012image,wang2015exposure,
  li2014selectively,sakai2015hybrid, nejati2017fast}.
  These fusion methods utilize a set of differently exposed images,
  ``multi-exposure images'', and fuse them to produce an image with high quality.
  Their development was inspired by high dynamic range (HDR) imaging techniques
  \cite{debevec1997recovering,reinhard2002photographic,oh2015robust,
  kinoshita2016remapping,kinoshita2017fast,kinoshita2017fast_trans,huo2016single,
  murofushi2013integer,murofushi2014integer,dobashi2014fixed}.
  The advantage of these methods, compared with HDR imaging techniques, is that
  they can eliminate three operations:
  generating HDR images, calibrating a camera response function (CRF),
  and preserving the exposure value of each photograph.

  However, the conventional multi-exposure image fusion methods have several problems
  due to the use of a set of differently exposed images.
  The set should consist of a properly exposed image,
  overexposed images and underexposed images,
  but determining appropriate exposure values is problematic.
  Moreover, even if appropriate exposure values are given,
  it is difficult to set them at the time of photographing.
  In particular, if the scene is dynamic or the camera moves while pictures
  are being captured, the exposure time should be shortened
  to prevent ghost-like or blurring artifacts in the fused image.
  
  To overcome these problems, this paper proposes a novel multi-exposure image
  fusion method based on exposure compensation.
  The goal of the proposed method is to produce appropriate multi-exposure images
  from input multi-exposure ones, and to generate an LDR image that clearly represents
  the overall image area, by fusing them.
  The proposed method adjusts the luminance of
  the input multi-exposure images on the basis of the relationship
  between the exposure values and pixel values,
  which is obtained by assuming that a digital camera
  has a linear response function.
  Moreover, the use of a local contrast enhancement method
  allows us to improve input multi-exposure images.
  The compensated images are finally combined by one of existing multi-exposure
  image fusion methods.
  
  We evaluate the effectiveness of the proposed method in terms of the
  quality of generated images.
  In the simulations, the proposed method is compared with conventional ones,
  by using the tone mapped image quality index (TMQI), statistical naturalness,
  and discrete entropy.
  The results show that the proposed method can produce images with higher quality
  than conventional ones.
\section{Preparation}
  Existing multi-exposure fusion methods use images taken under different
  exposure conditions, i.e., ``multi-exposure images.''
  Here we discuss the relationship between exposure values and pixel values.
  For simplicity, we focus on grayscale images in this section.
\subsection{Relationship between exposure values and pixel values}
\begin{figure}[!t]
    \centering
    \includegraphics[width=0.95\linewidth]{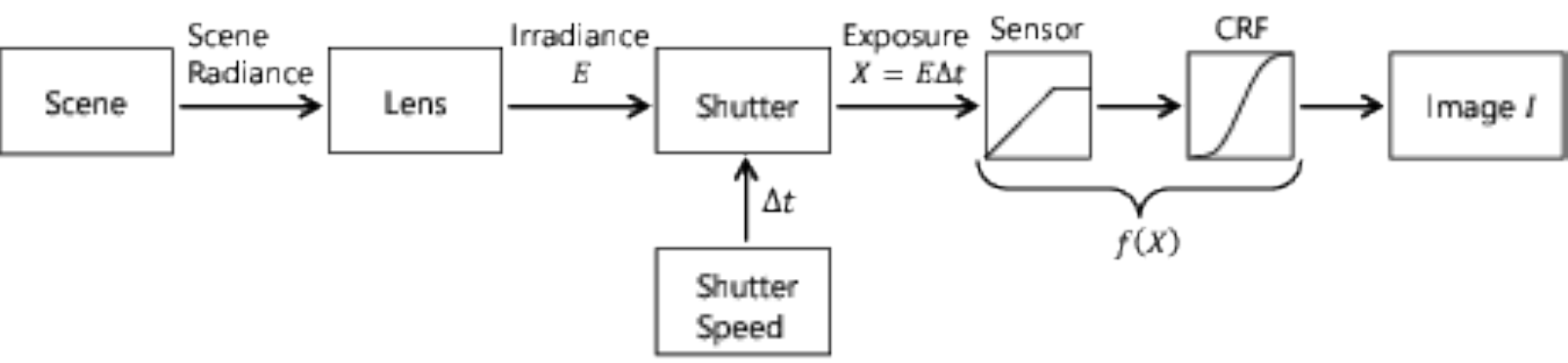}
    \caption{Imaging pipeline of digital camera \label{fig:camera}}
  \end{figure}

  Figure \ref{fig:camera} shows a typical imaging pipeline for
  a digital camera\cite{dufaux2016high}.
  The radiant power density at the sensor, i.e., irradiance $E$,
  is integrated over the time $\Delta t$ the shutter is open,
  producing an energy density, commonly referred to as exposure $X$.
  If the scene is static during this integration,
  exposure $X$ can be written simply as the product of irradiance $E$ and
  integration time $\Delta t$ (referred to as ``shutter speed''):
  \begin{equation}
    X(p) = E(p)\Delta t,
    \label{eq:exposure}
  \end{equation}
  where $p=(x,y)$ indicates the pixel at point $(x,y)$.
  A pixel value $I(p) \in [0, 1]$ in the output image $I$ is given by
  \begin{equation}
    I(p) = f(X(p)),
    \label{eq:CRF}
  \end{equation}
  where $f$ is a function combining sensor saturation
  and a camera response function (CRF).
  The CRF represents the processing in each camera which makes the final image $I(p)$
  look better.

  Camera parameters, such as shutter speed and lens aperture,
  are usually calibrated in terms of exposure value (EV) units,
  and the proper exposure for a scene is automatically decided by
  the camera.
  The exposure value is commonly controlled by changing the shutter speed
  although it can also be controlled by adjusting various camera parameters.
  Here we assume that the camera parameters except for
  the shutter speed are fixed.
  Let $\nu = 0 \mathrm{[EV]}$ and $\Delta \tau$
  be the proper exposure value
  and shutter speed under the given conditions, respectively.
  The exposure value $v_i \mathrm{[EV]}$ of an image taken at
  shutter speed $\Delta t_i$ is given by
  \begin{equation}
    v_i = \log_2 \Delta t_i - \log_2 \Delta \tau.
    \label{eq:EV}
  \end{equation}
  From (\ref{eq:exposure}) to (\ref{eq:EV}),
  images $I_0$ and $I_i$ exposed at $0 \mathrm{[EV]}$ and $v_i \mathrm{[EV]}$,
  respectively, are written as
  \begin{align}
    I_0(p) &= f(E(p)\Delta \tau) \label{eq:CRFwithExposure}\\
    I_i(p) &= f(E(p)\Delta t_i) \label{eq:CRFwithExposure2}
            = f(2^{v_i} E(p)\Delta \tau).
  \end{align}
  Assuming function $f$ is linear,
  we obtain the following relationship between $I_0$ and $I_i$:
  \begin{equation}
    I_i(p) = 2^{v_i} I_0(p).
    \label{eq:relationship}
  \end{equation}
  Therefore, the exposure can be varied artificially by multiplying $I_0$ by a constant.
  This ability is used in a new multi-exposure fusion method,
  which is described in the next section.

\subsection{Scenario}
  For a multi-exposure fusion method to produce high quality images,
  the input images should represent the bright, middle, and dark
  regions of the scene.
  These images generally consist of
  a properly exposed image ($v_i = 0 \mathrm{[EV]}$),
  overexposed images ($v_i > 0$), and underexposed images ($v_i < 0$).
  For example, three multi-exposure images might be taken at $v_i = -1, 0, +1 \mathrm{[EV]}$.

  However, there are several problems in photographing multi-exposure images.
  \begin{itemize}[nosep]
    \item Determining appropriate exposure values
      for multi-exposure image fusion.
    \item Setting appropriate exposure values at the time of photographing
      when there are time constraints.
    \item Using an image taken at $0 \mathrm{[EV]}$ as it
      might not represent the scene properly.
  \end{itemize}
  To overcome these problems, this paper proposes a novel multi-exposure fusion method
  based on the relationship between the exposure values and pixel values.
\section{Proposed multi-exposure image fusion}
\begin{figure*}[!t]
    \centering
    \includegraphics[clip, width=12cm]{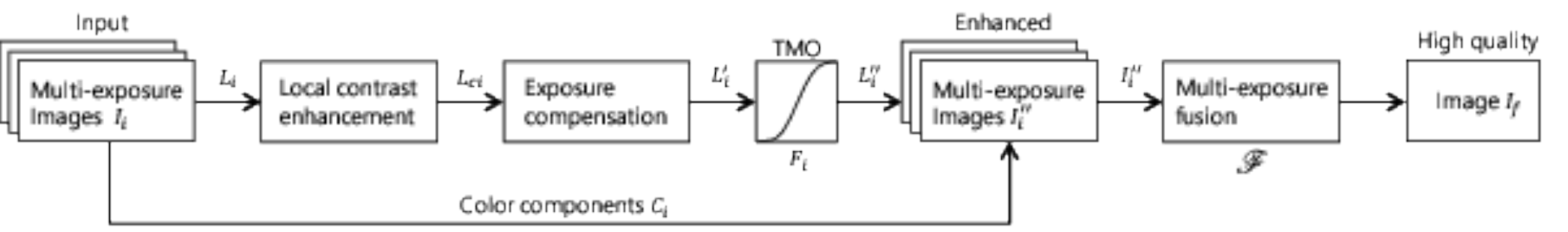}
    \caption{Outline of proposed method \label{fig:proposedMEF}}
  \end{figure*}

  The outline of the proposed method is illustrated in Fig. \ref{fig:proposedMEF}.
  To enhance the quality of multi-exposure images, local contrast enhancement is applied to
  luminance $L_i (1 \le i \le N, i \in \mathbb{N})$
  calculated from the $i$-th input image $I_i$, and then
  exposure compensation and tone mapping are applied.
  Next, image $I_f$ with improved quality is produced by existing multi-exposure image fusion.
  Here we consider input image $I_i$ with exposure value $v_i$
  that satisfies $v_i < v_{i+1}$.
\subsection{Local contrast enhancement}
  If the input images do not represent the scene clearly,
  the quality of an image fused from them will be lower than
  that of an image fused from ideally exposed images.
  Therefore, the dodging and burning algorithm is used to enhance
  the local contrast\cite{huo2013dodging}.
  The luminance $L_{ci}$ enhanced by the algorithm is given by
  \begin{equation}
    L_{ci}(p) = \frac{L_i^2(p)}{L_{ai}(p)},
    \label{eq:dodgingAndBurning}
  \end{equation}
  where $L_{ai}(p)$ is the local average of luminance $L_i(p)$ around pixel $p$.
  It is obtained by applying a low-pass filter to $L_i(p)$.
  Here, a bilateral filter is used for this purpose.

  $L_{ai}(p)$ is calculated using the bilateral filter:
  \begin{equation}
    L_{ai}(p) = \frac{1}{c_i(p)}
              \sum_{q \in \Omega}
                L_i(q) g_{\sigma_1}(q-p) g_{\sigma_2}(L_i(q) - L_i(p)),
    \label{eq:bilateral}
  \end{equation}
  where $\Omega$ is the set of all pixels, and $c_i(p)$ is a normalization term such as
  \begin{equation}
    c_i(p) = \sum_{q \in \Omega} g_{\sigma_1}(q-p) g_{\sigma_2}(L_i(q) - L_i(p)),
    \label{eq:normalizingConst}
  \end{equation}
  where $g_{\sigma}$ is a Gaussian function given by
  \begin{equation}
    g_{\sigma}(p | p=(x,y)) = C_{\sigma}\exp \left( -\frac{x^2 + y^2}{\sigma^2} \right)
    \label{eq:gaussian}
  \end{equation}
  using a normalization factor $C_{\sigma}$.
  Parameters $\sigma_1 = 16$ and $\sigma_2 = 3/255$ are set
  in accordance with \cite{huo2013dodging}.
\subsection{Exposure compensation}
  The purpose of the exposure compensation is to adjust the luminance of
  each input image $I_i$, so that adjusted images have appropriate exposure values
  for multi-exposure image fusion.
  The luminance $L'_i$ of adjusted image $I'_i$ is simply obtained by,
  according to eq. (\ref{eq:relationship}),
  \begin{equation}
    L'_i(p) = \alpha_i L_{ci}(p),
    \label{eq:constMultiplication}
  \end{equation}
  where parameter $\alpha_i > 0$ indicates the degree of adjustment.
  Next, the way to estimate the parameter $\alpha_i$ is described.
\subsubsection{Estimating parameter $\alpha_i$}
  In $N$ input images, the $j = \lceil \frac{N + 1}{2} \rceil$-th image $I_j$ has middle
  brightness, and the overexposed (or underexposed) areas in $I_j$ are smaller than
  those in the other images. Therefore, the quality of image $I_j$
  should be better than that of the other images.
  We thus estimate parameter $\alpha_j$ from the $j$-th image in order to
  map the geometric mean $\overline{L}_{cj}$ of luminance $L_{cj}$ to middle-gray
  of the displayed image, or 0.18 on a scale from zero to one,
  as in \cite{reinhard2002photographic},
  where the geometric mean of the luminance values indicates
  the approximate brightness of the image.
  
  The geometric mean $\overline{L}_{ci}$ of luminance $L_{ci}(p)$ is calculated using 
  \begin{equation}
    \overline{L}_{ci} =
      \exp{
        \left(\frac{1}{|\Omega|}
          \sum_{p \in \Omega} \log{\left(\max{\left( L_{ci}(p), \epsilon \right)}\right)}
        \right)
      },
    \label{eq:geoMeanEps}
  \end{equation}
  where $\epsilon$ is set to a small value to avoid singularities at $L_{ci}(p)=0$.
  Parameter $\alpha_j$ is derived using eq. (\ref{eq:geoMeanEps}) from
  \begin{equation}
    \alpha_j = \frac{0.18}{\overline{L}_{cj}}.
    \label{eq:alphaj}
  \end{equation}
  The adjusted version $I'_k$ of the $k$-th input image $I_k (k \neq j)$ should be
  brighter or darker than $I'_j$ if $k > j$ or $k < j$, respectively.
  Therefore, parameter $\alpha_k$ is calculated as
  \begin{equation}
    \alpha_k = \frac{0.18\cdot2^{k-j}}{\overline{L}_{ck}}.
    \label{eq:unknownEV}
  \end{equation}
\subsection{Tone mapping}
  Since the adjusted luminance value $L'_i(p)$ often exceeds
  the maximum value of the common image format,
  pixel values might be lost due to truncation of the values.
  This problem is overcome by using a tone mapping operation
  to fit the adjusted luminance value into the interval $[0, 1]$.

  The luminance $L''_i$ of an enhanced multi-exposure image is obtained
  by applying a tone mapping operator $F_i$ to $L'_i$:
  \begin{equation}
    L''_i(p) = F_i(L'_i(p)).
    \label{eq:TM}
  \end{equation}
  Reinhard's global operator is used here as a tone mapping operator $F_i$
  \cite{reinhard2002photographic}.
  
  Reinhard's global operator is given by
  \begin{equation}
    F_i(L(p)) = \frac{L(p)\left(1 + \frac{L(p)}{L^2_{white_i}} \right)}{1 + L(p)},
    \label{eq:reinhardTMO}
  \end{equation}
  where parameter $L_{white_i} > 0$ determines luminance value $L(p)$
  as $F_i(L(p)) = 1$.
  Note that Reinhard's global operator $F_i$ is a monotonically increasing function.
  Here, let $L_{white_i} = \max L'_i(p)$.
  We obtain $L''_i(p) \le 1$ for all $p$.
  Therefore, truncation of the luminance values can be prevented.

  Combining $L''_i$,
  luminance $L_i$ of the $i$-th input image $I_i$,
  and RGB pixel values $C_i(p) \in \{R_i(p), G_i(p), B_i(p)\}$ of $I_i$,
  we obtain RGB pixel values $C''_i(p) \in \{R''_i(p), G''_i(p), B''_i(p)\}$ of
  the enhanced multi-exposure images $I''_i$:
  \begin{equation}
    C''_i(p) = \frac{L''_i(p)}{L_i(p)}C_i(p).
    \label{eq:color}
  \end{equation}
\subsection{Fusion of enhanced multi-exposure images}
  Enhanced multi-exposure images $I''_i$ can be used as input for any existing
  multi-exposure image fusion methods.
  While numerous methods for fusing images have been proposed,
  here we use those of Mertens et al.\cite{mertens2009exposure},
  Sakai et al.\cite{sakai2015hybrid},
  and Nejati et al.\cite{nejati2017fast}.
  A final image $I_f$ is produced as
  \begin{equation}
    I_f = \mathscr{F}(I''_1, I''_2, \cdots, I''_N),
    \label{eq:fusion}
  \end{equation}
  where $\mathscr{F}(I_1, I_2, \cdots, I_N)$ indicates a function to fuse $N$ images
  $I_1, I_2, \cdots, I_N$ into a single image.
\section{Simulation}
  We evaluated the proposed method in terms of the quality of generated images $I_f$.
\subsection{Comparison with conventional methods}
  Two simulations ``Simulation 1'' and ``Simulation 2'' were carried out
  to evaluate the effectiveness of the proposed method.
  Three fusion methods mentioned above were used as a fusion method $\mathscr{F}$.
  
  To evaluate the quality of the images produced by each method,
  objective quality assessments are needed.
  Typical quality assessments such as the peak signal to noise ratio (PSNR)
  and the structural similarity index (SSIM) are not suitable for this purpose
  because they use the target image with the highest quality as the reference one.
  We therefore used the tone mapped image quality index (TMQI) \cite{yeganeh2013objective}
  and discrete entropy as quality assessments as the quality assessments
  as they do not require a reference image.

  TMQI represents the quality of an image tone mapped from an HDR image;
  the index incorporates structural fidelity and statistical naturalness.
  An HDR image is used as a reference to calculate structural fidelity.
  A reference is not needed to calculate statistical naturalness.
  Since the processes of tone mapping and photographing are similar,
  TMQI is also useful for evaluating photographs.
  Discrete entropy represents the amount of information in an image.
\subsection{Simulation conditions}
  The conditions of two simulations are described here.
  The differences between the two simulations is in how to prepare input images,
  i.e., original ones.
  \subsubsection{Simulation 1 (using HDR images)}
  In Simulation 1,
  HDR images were utilized to prepare input images.
  The following procedure was used to evaluate the effectiveness of the proposed method.
  \begin{enumerate}[nosep]
    \item Map HDR image $I_H$ to three multi-exposure images $I_i, i = 1,2,3$
      with exposure values $v_i=i-2\mathrm{[EV]}$
      by using a tone mapping operator (see Fig. \ref{fig:inputImages}\subref{fig:OrgP1EV}).
    \item Obtain $I_f$ from $I_i$ using the proposed method.
    \item Compute TMQI values between $I_f$ and $I_H$.
    \item Compute discrete entropy of $I_f$.
  \end{enumerate}
  In step 1, the tone mapping operator corresponds to function $f$
  in eqs. (\ref{eq:CRFwithExposure}) and (\ref{eq:CRFwithExposure2}) (see Fig. \ref{fig:camera}).
  As assumed for eq. (\ref{eq:relationship}),
  a linear operator was utilized as the tone mapping operator.

  We used 60 HDR images selected from available online
  databases\cite{openexrimage,anyherehdrimage}.
  \subsubsection{Simulation 2 (photographing directly)}
\begin{figure}[!t]
  \centering
  \subfloat[GoldenGate ($v_i=-1, 0, +1\mathrm{[EV]}$)]{
    \includegraphics[width=0.32\hsize]{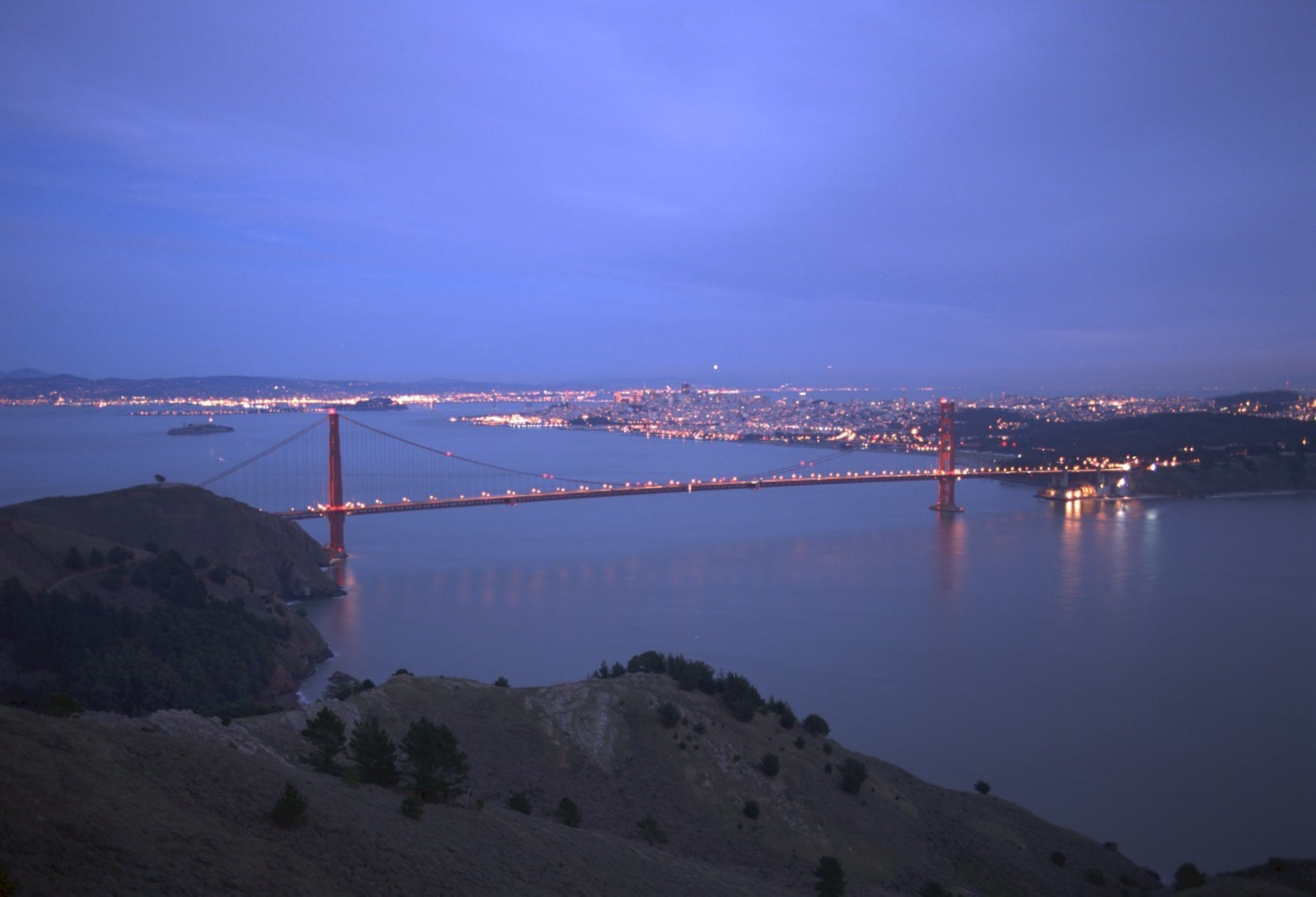}
    \includegraphics[width=0.32\hsize]{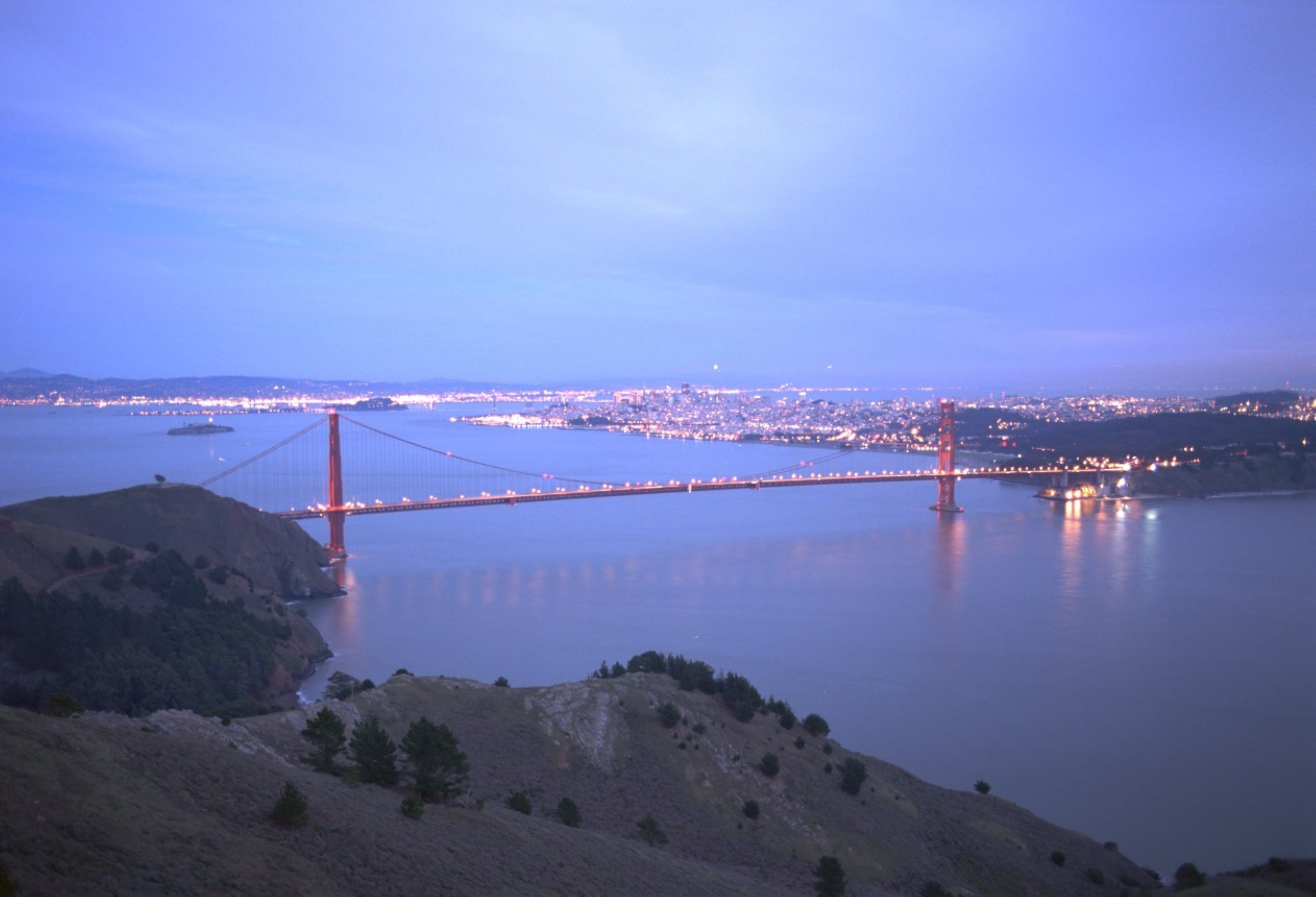}
    \includegraphics[width=0.32\hsize]{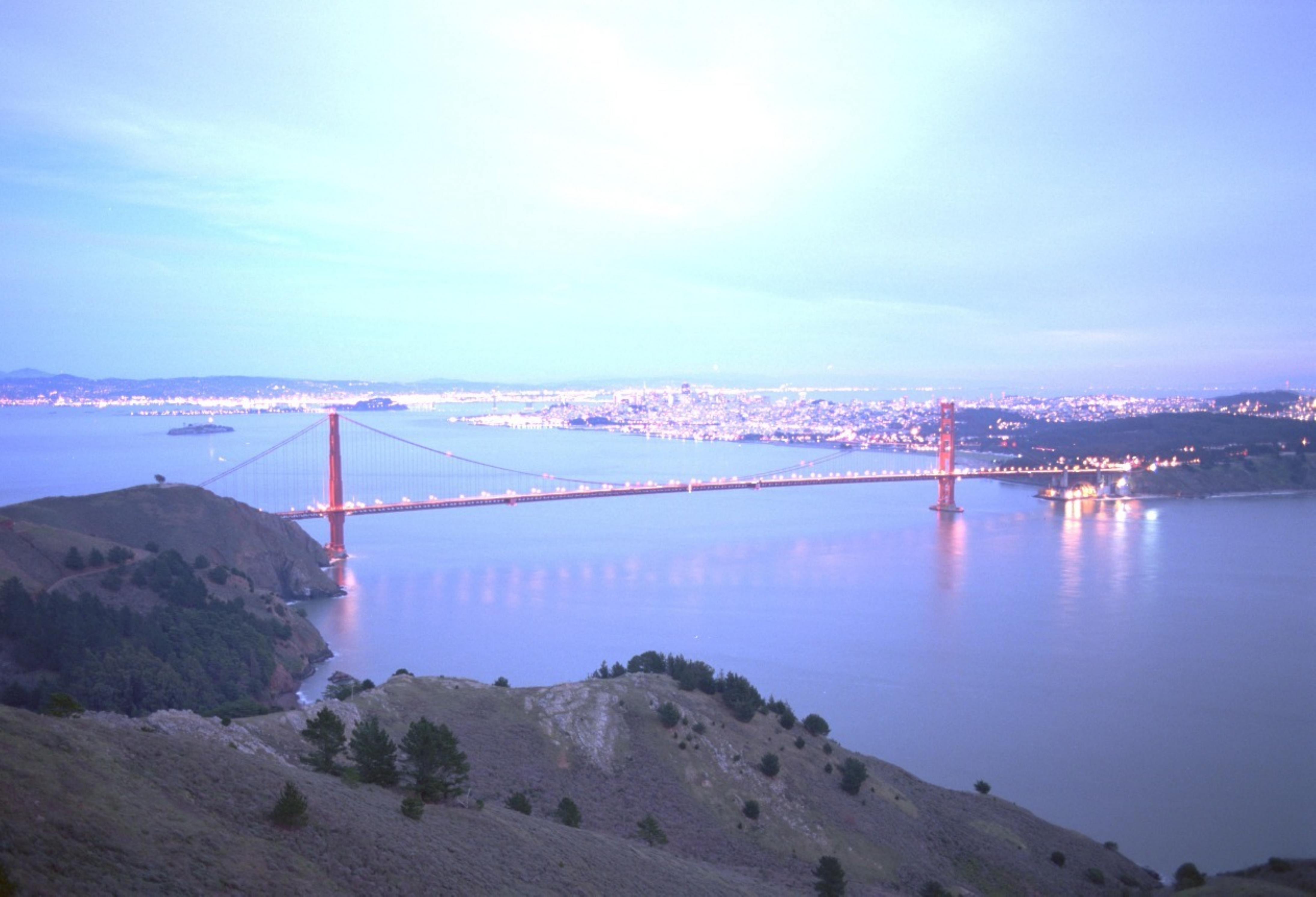}
    \label{fig:OrgP1EV}}\\
  \subfloat[Lobby ($v_i=-1, 0, +1\mathrm{[EV]}$)]{
    \includegraphics[width=0.32\hsize]{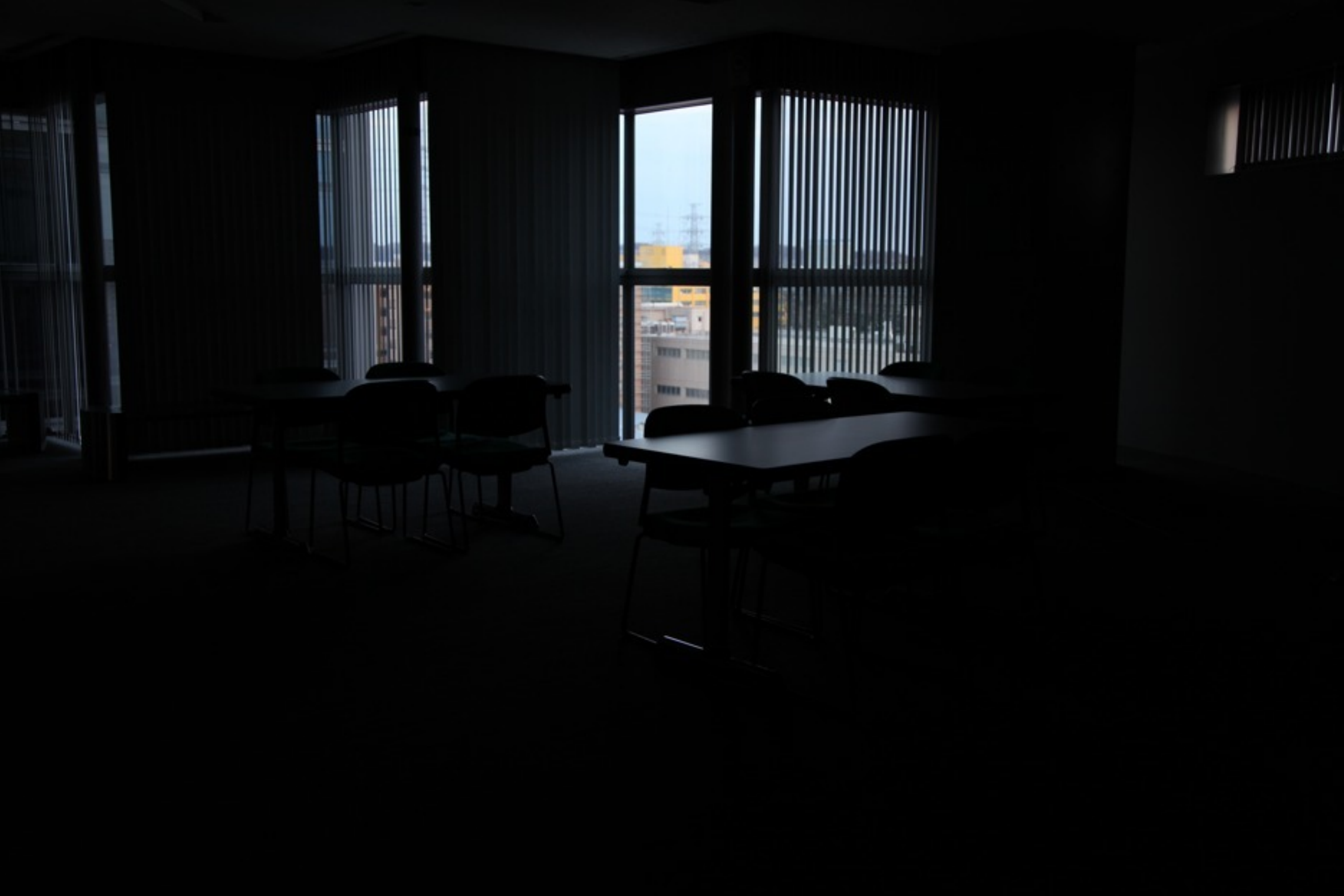}
    \includegraphics[width=0.32\hsize]{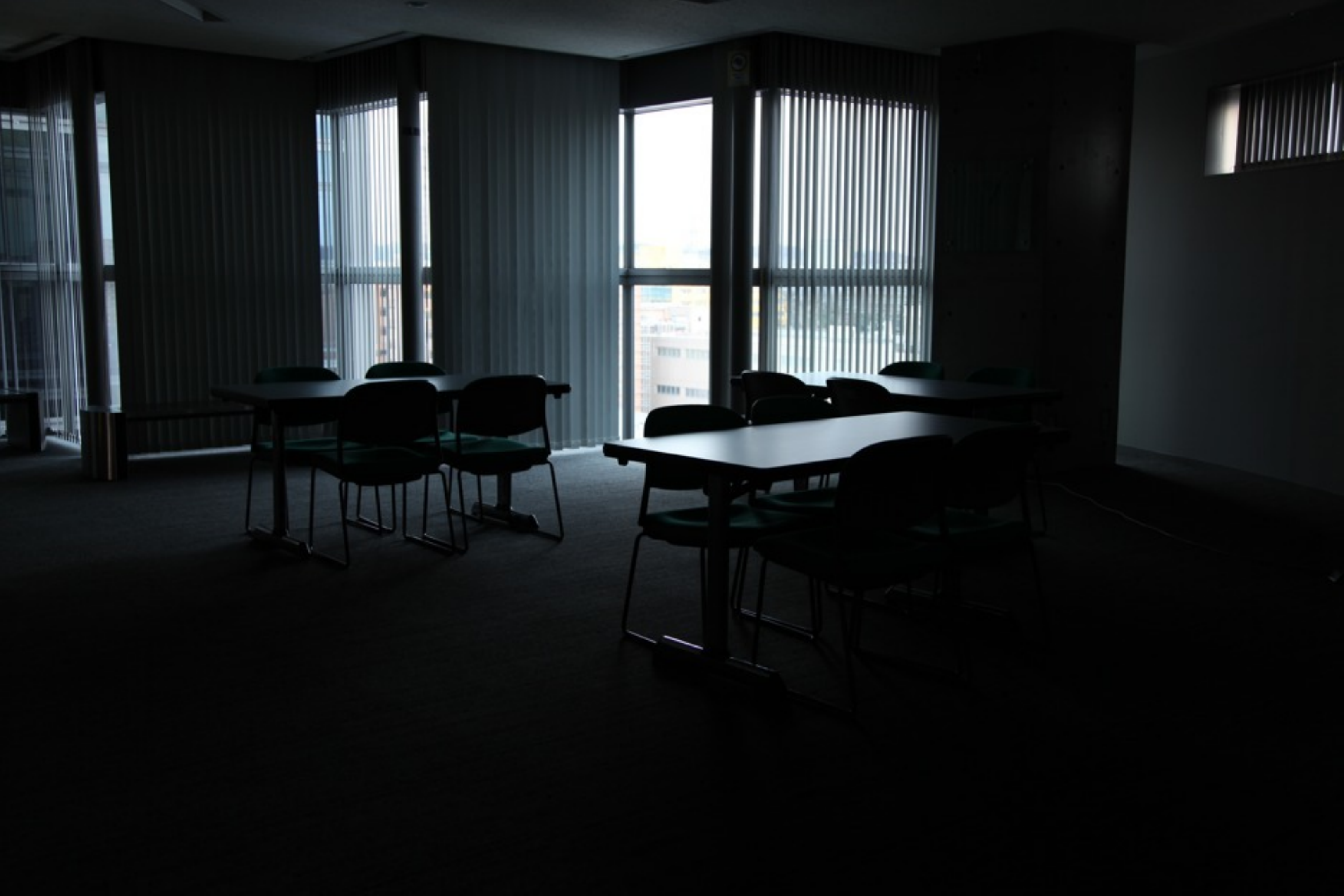}
    \includegraphics[width=0.32\hsize]{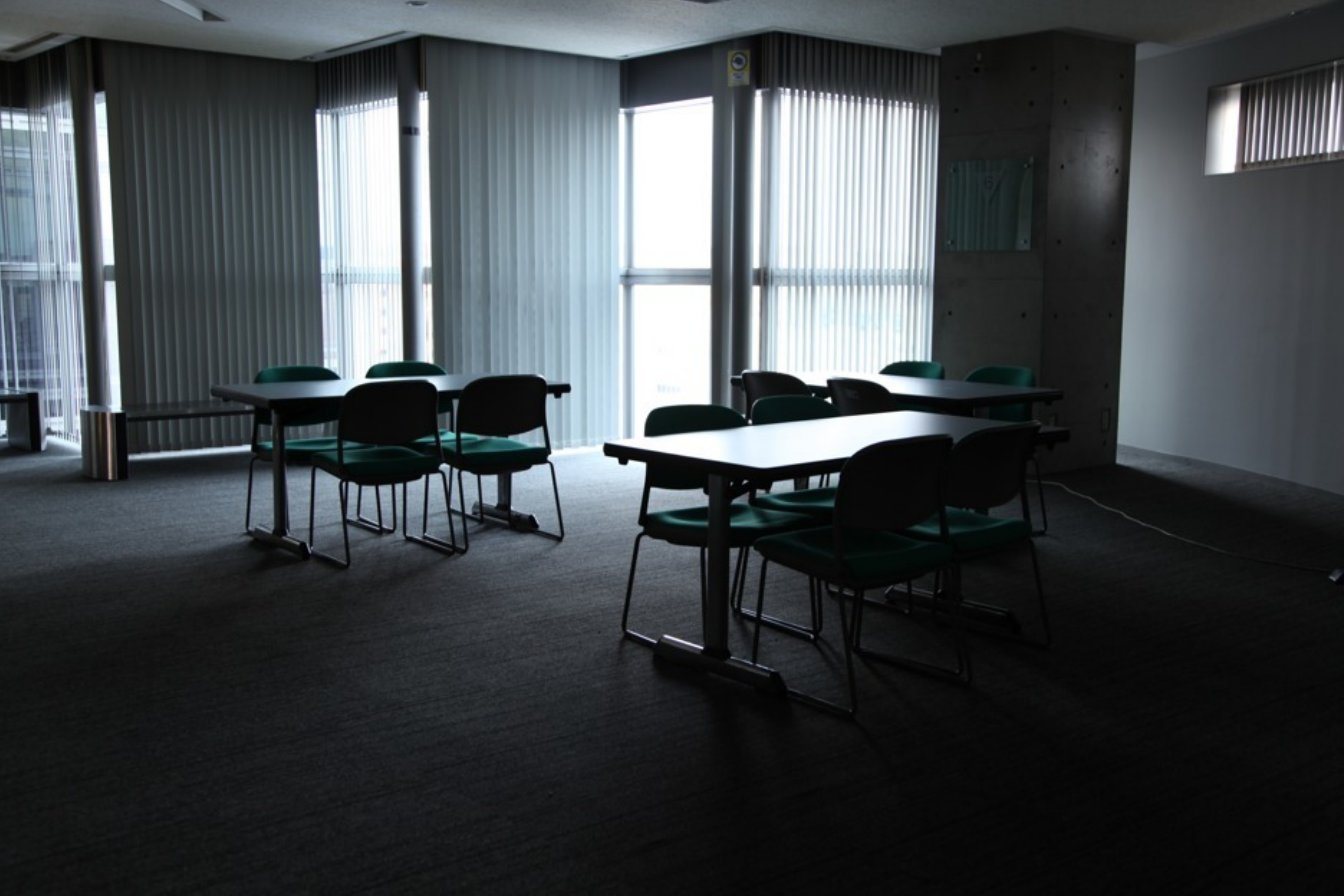}
    \label{fig:LobbyP1EV}}\\
  \subfloat[Lobby (enhanced by the proposed method)]{
    \includegraphics[width=0.32\hsize]{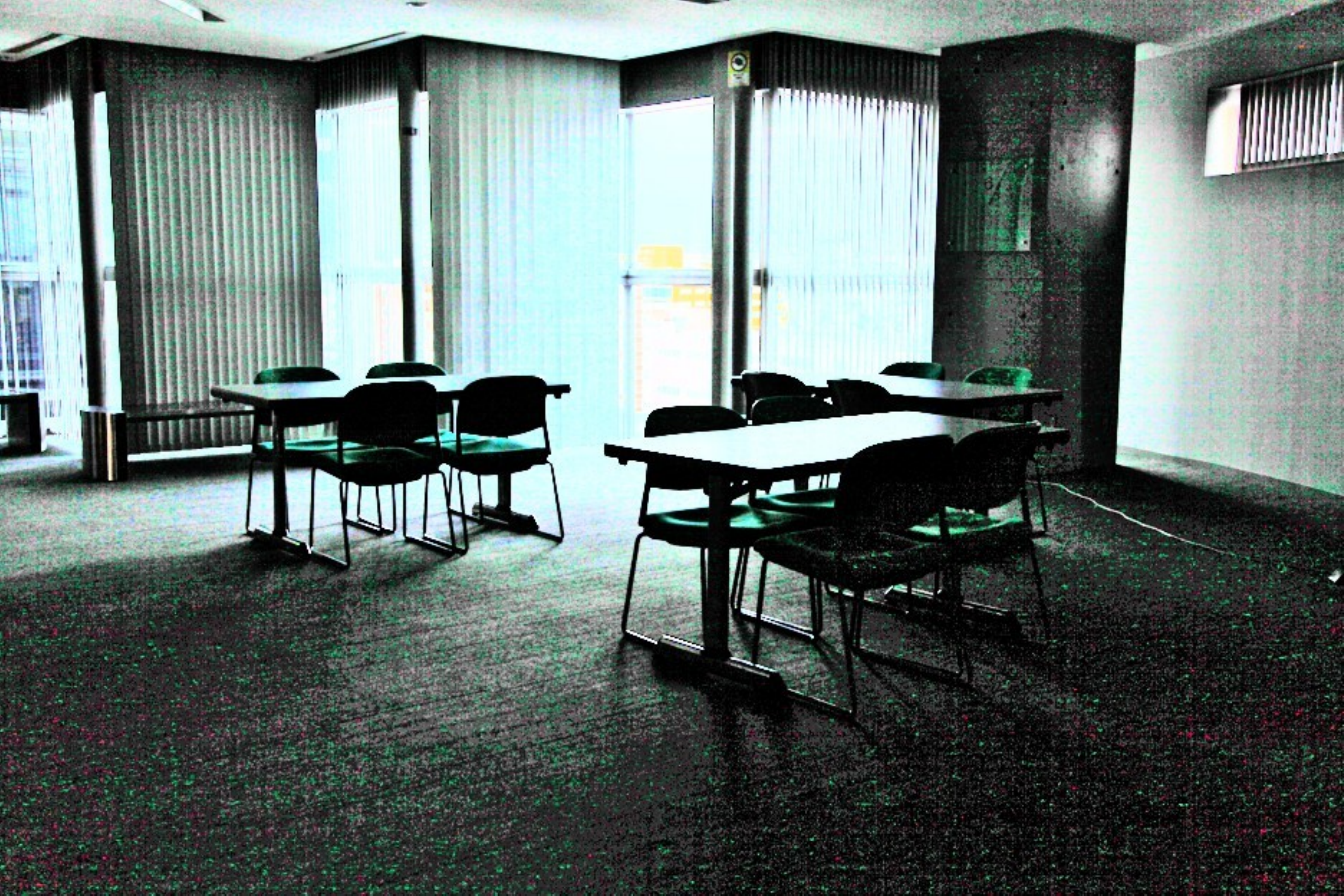}
    \includegraphics[width=0.32\hsize]{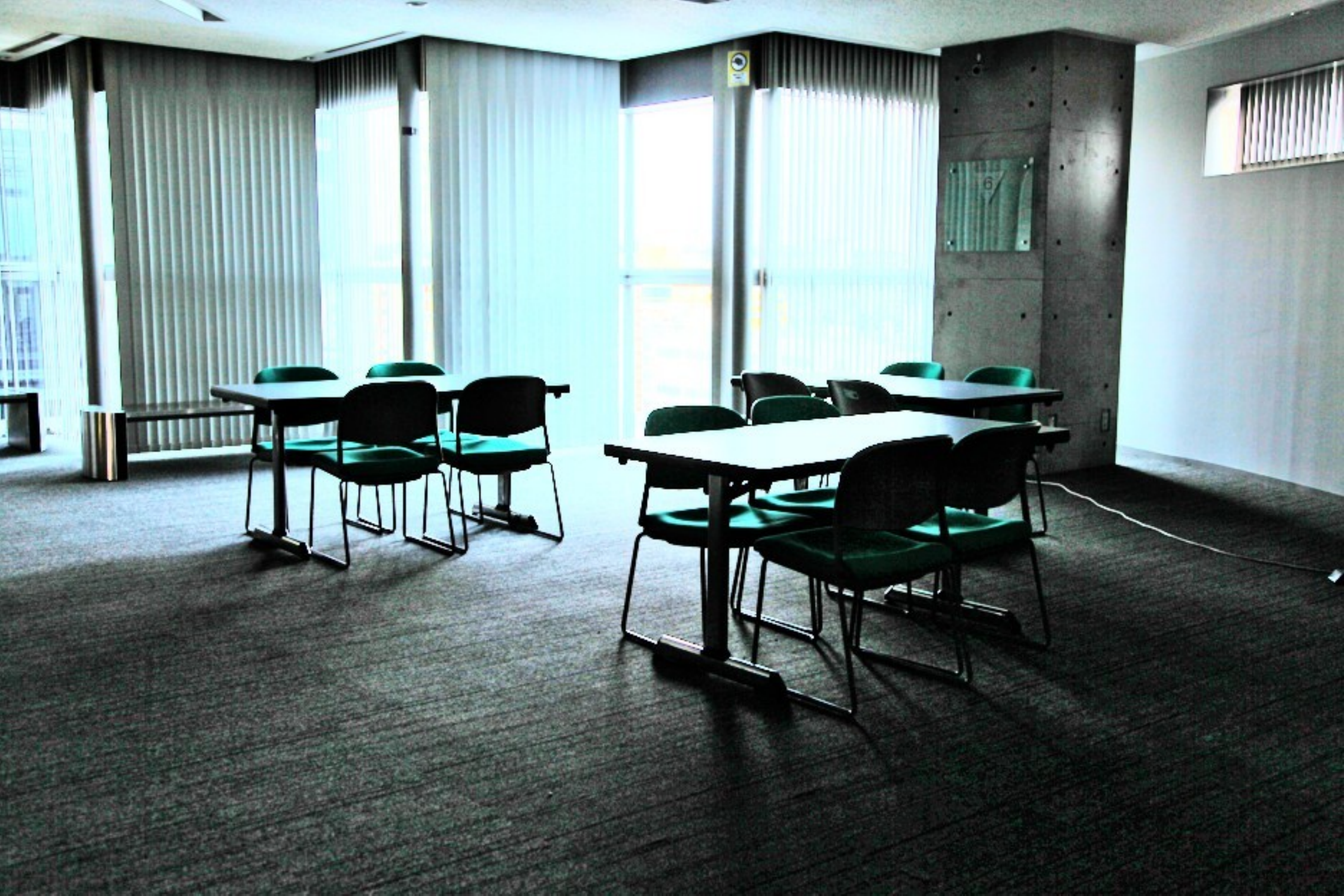}
    \includegraphics[width=0.32\hsize]{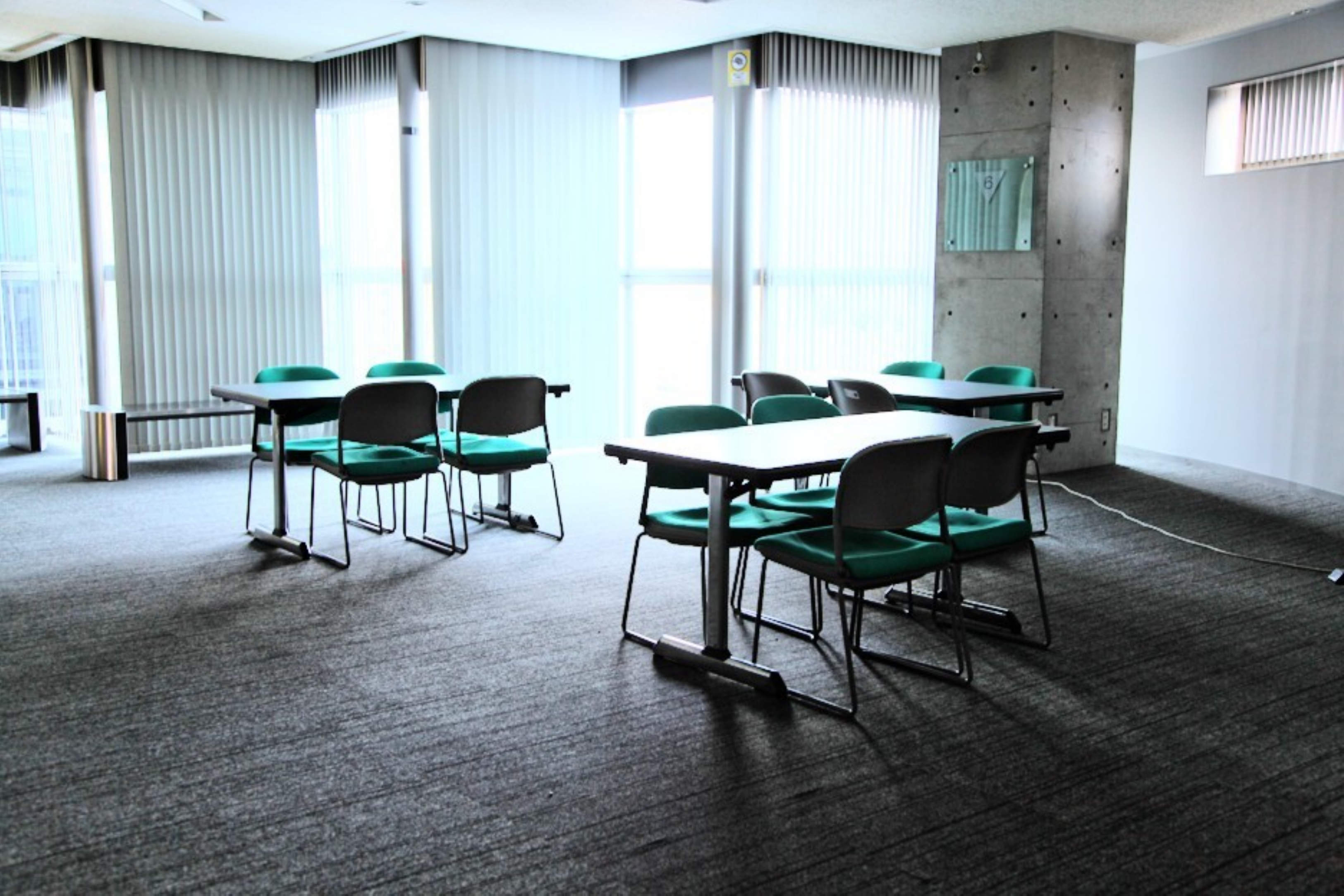}
    \label{fig:EnhancedLobby}}\\
  \caption{Examples of input multi-exposure images $I_i$}
  \label{fig:inputImages}
\end{figure}

  In Simulation 2,
  photographs taken by Canon EOS 5D Mark II camera were directly used as input images $I_i$
  (see Fig. \ref{fig:inputImages}\subref{fig:LobbyP1EV}).
  Since there were no HDR images for Simulation 2,
  the first step in Simulation 1 was not needed.
  In addition, structural fidelity in the TMQI could not be calculated due to the non-use
  of HDR images.
  Thus, we used only statistical naturalness in the TMQI for the evaluation.
  Other conditions were the same as for Simulation 1.
\subsection{Simulation results}
  Simulation results are illustrated under Simulation 1 and Simulation 2, respectively,
  where a higher score indicates better visual quality.
\subsubsection{Simulation 1}
\begin{figure}[!t]
  \centering
  \subfloat{
    \includegraphics[width=0.32\hsize]{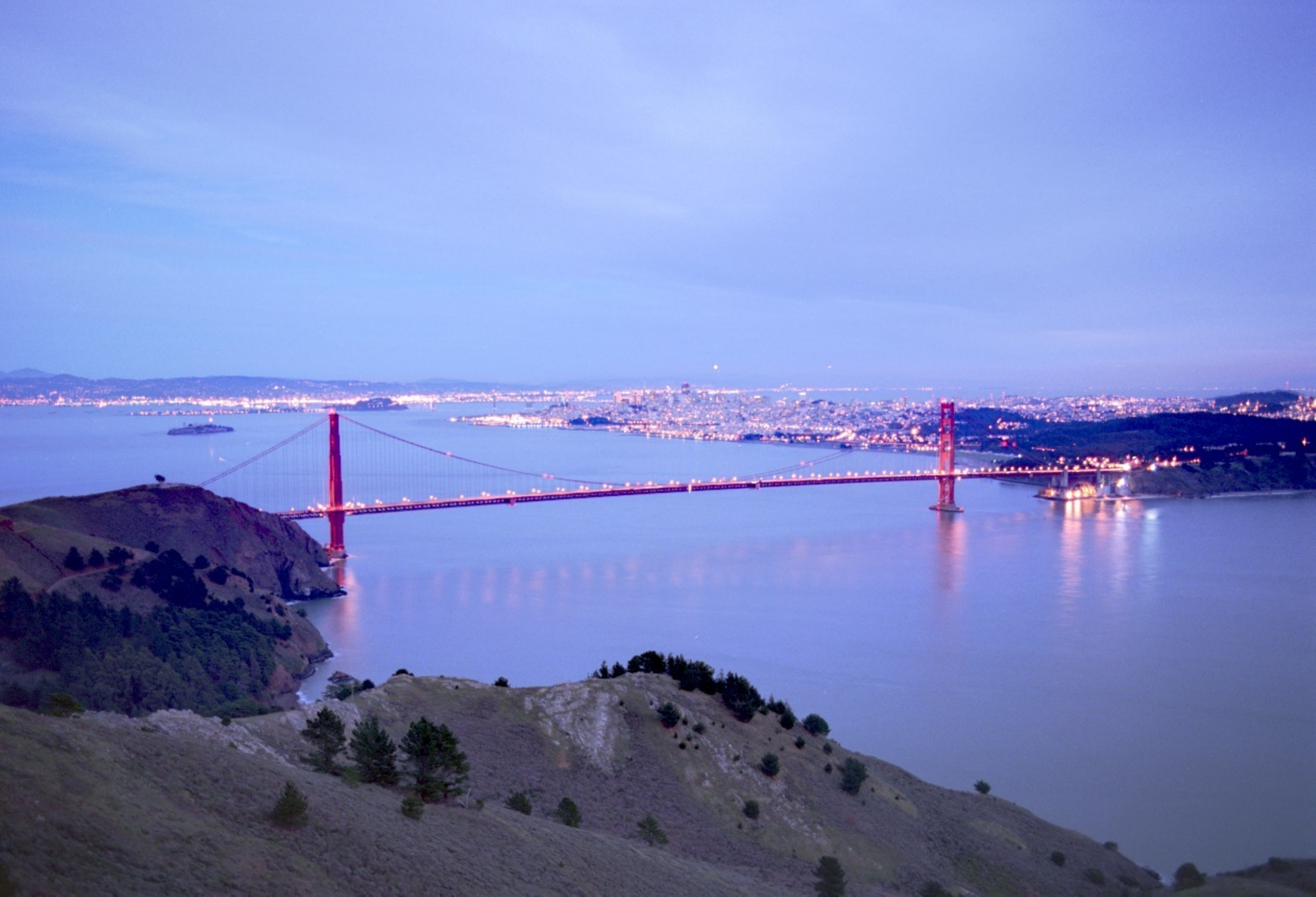}
    \label{fig:Mertens}}
  \subfloat{
    \includegraphics[width=0.32\hsize]{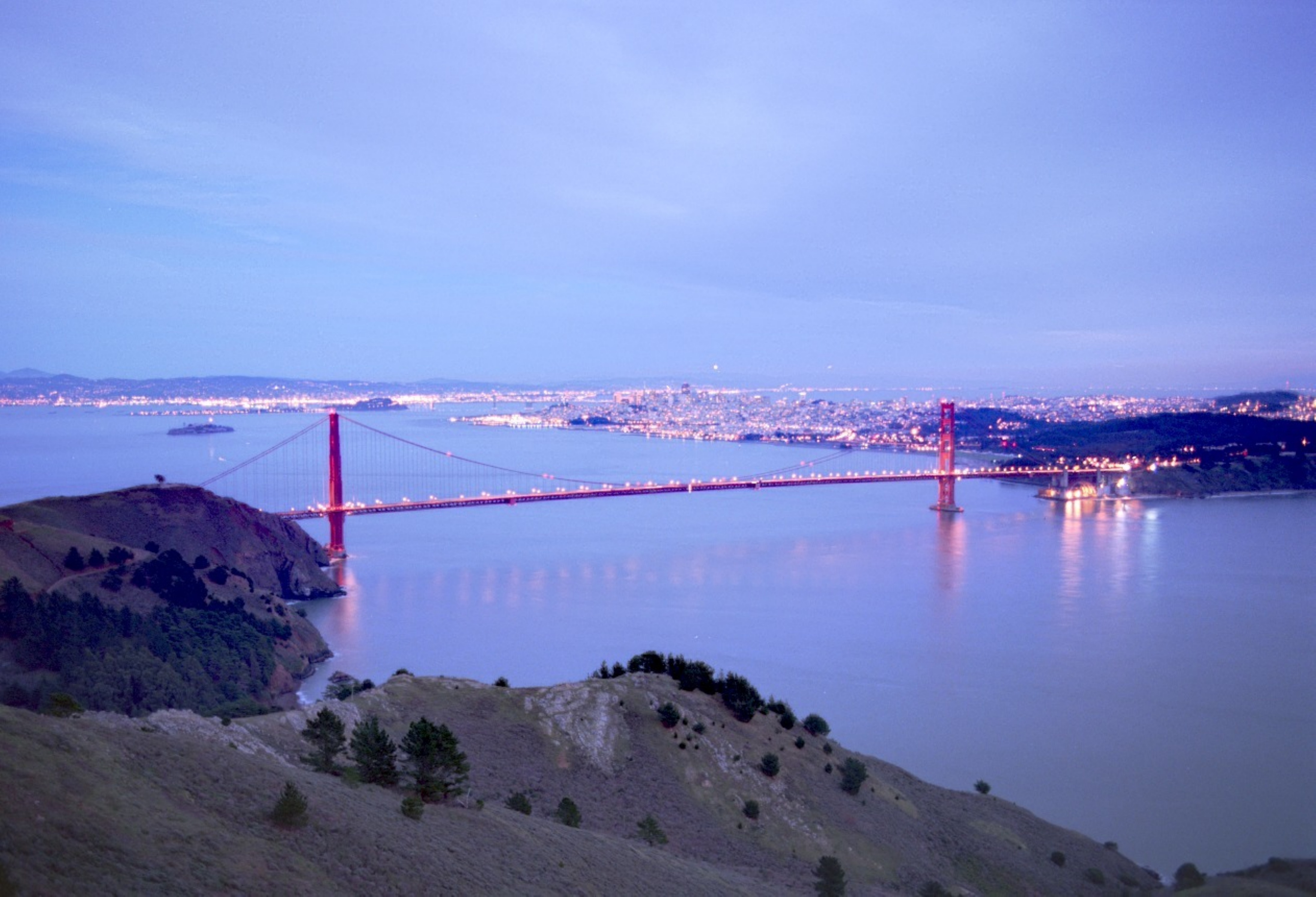}
    \label{fig:Yoshida}}
  \subfloat{
    \includegraphics[width=0.32\hsize]{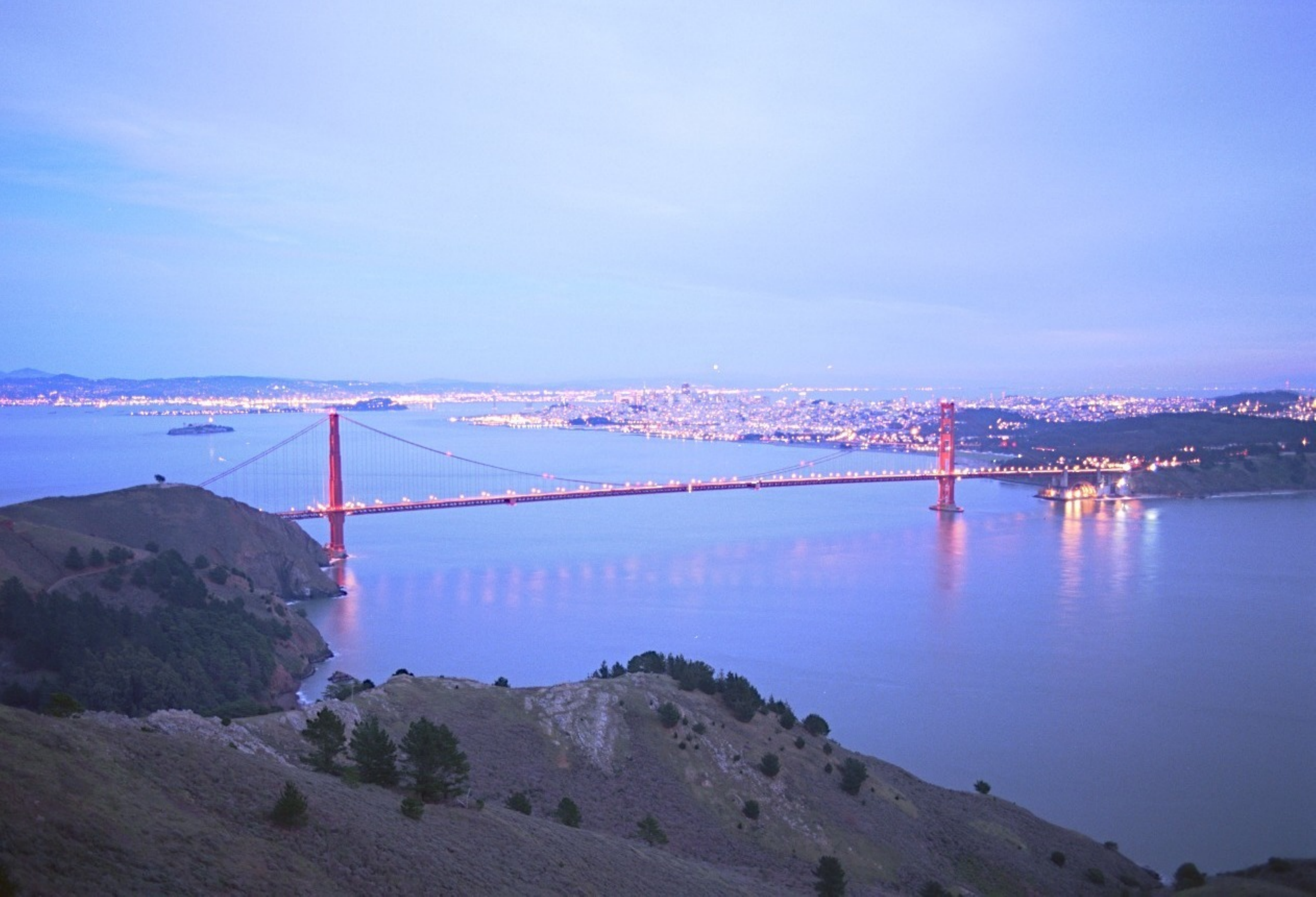}
    \label{fig:Nejati}}\\
  \vspace{2pt}
  \addtocounter{subfigure}{-3}
  \subfloat[Mertens\cite{mertens2009exposure}]{
    \includegraphics[width=0.32\hsize]{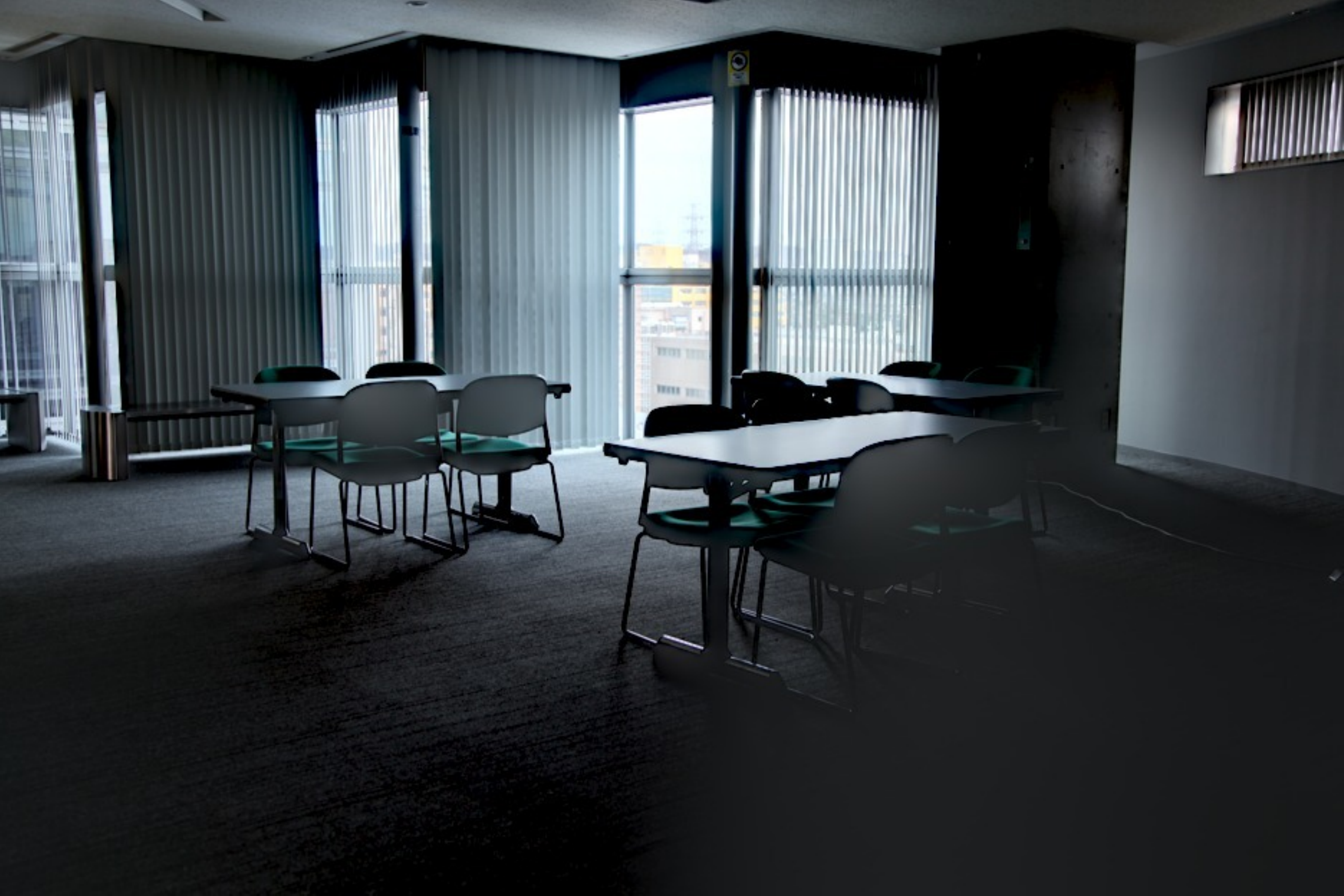}
    \label{fig:Lobby_Mertens}}
  \subfloat[Sakai\cite{sakai2015hybrid}]{
    \includegraphics[width=0.32\hsize]{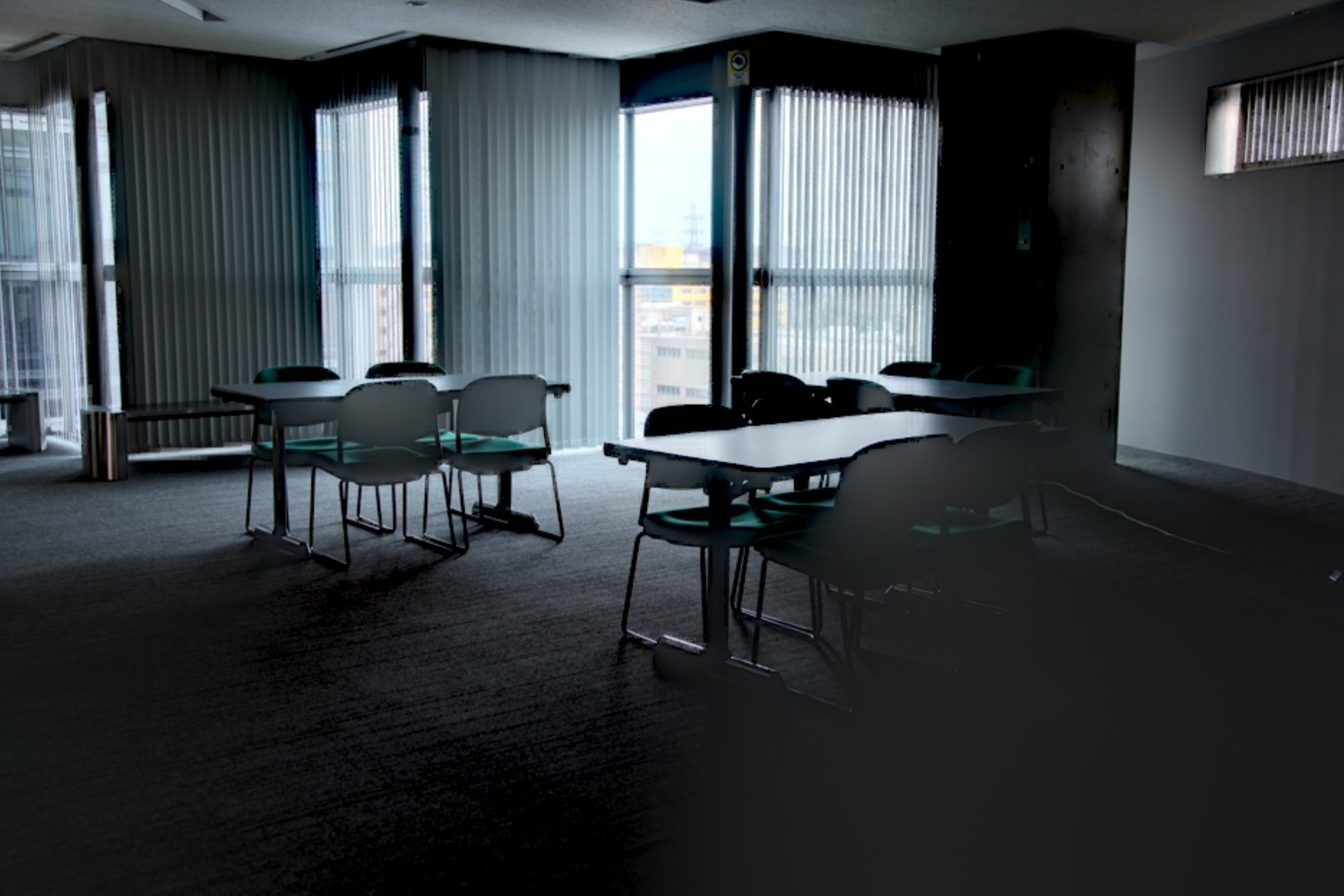}
    \label{fig:Lobby_Yoshida}}
  \subfloat[Nejati\cite{nejati2017fast}]{
    \includegraphics[width=0.32\hsize]{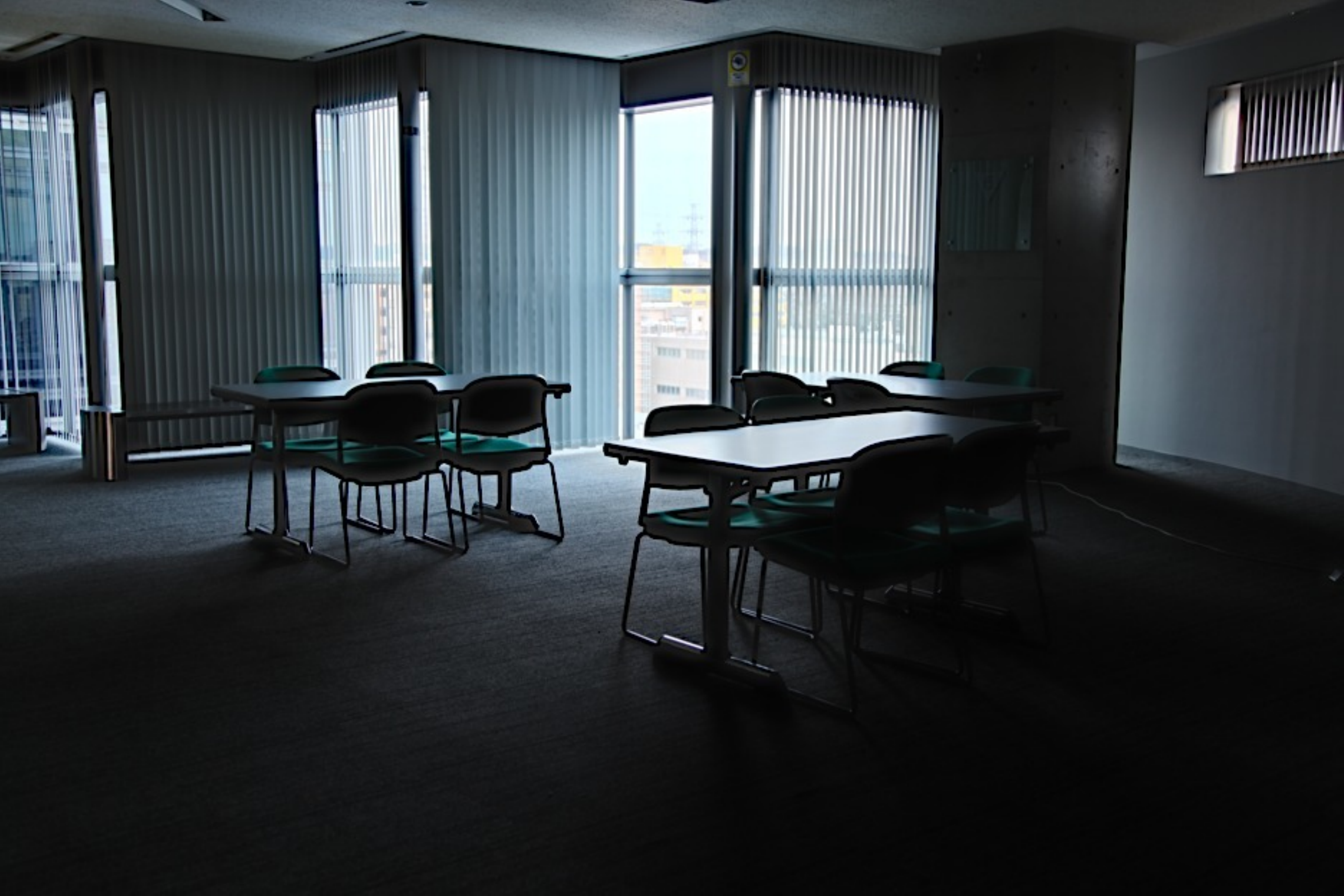}
    \label{fig:Lobby_Nejati}}\\
  \subfloat{
    \includegraphics[width=0.32\hsize]{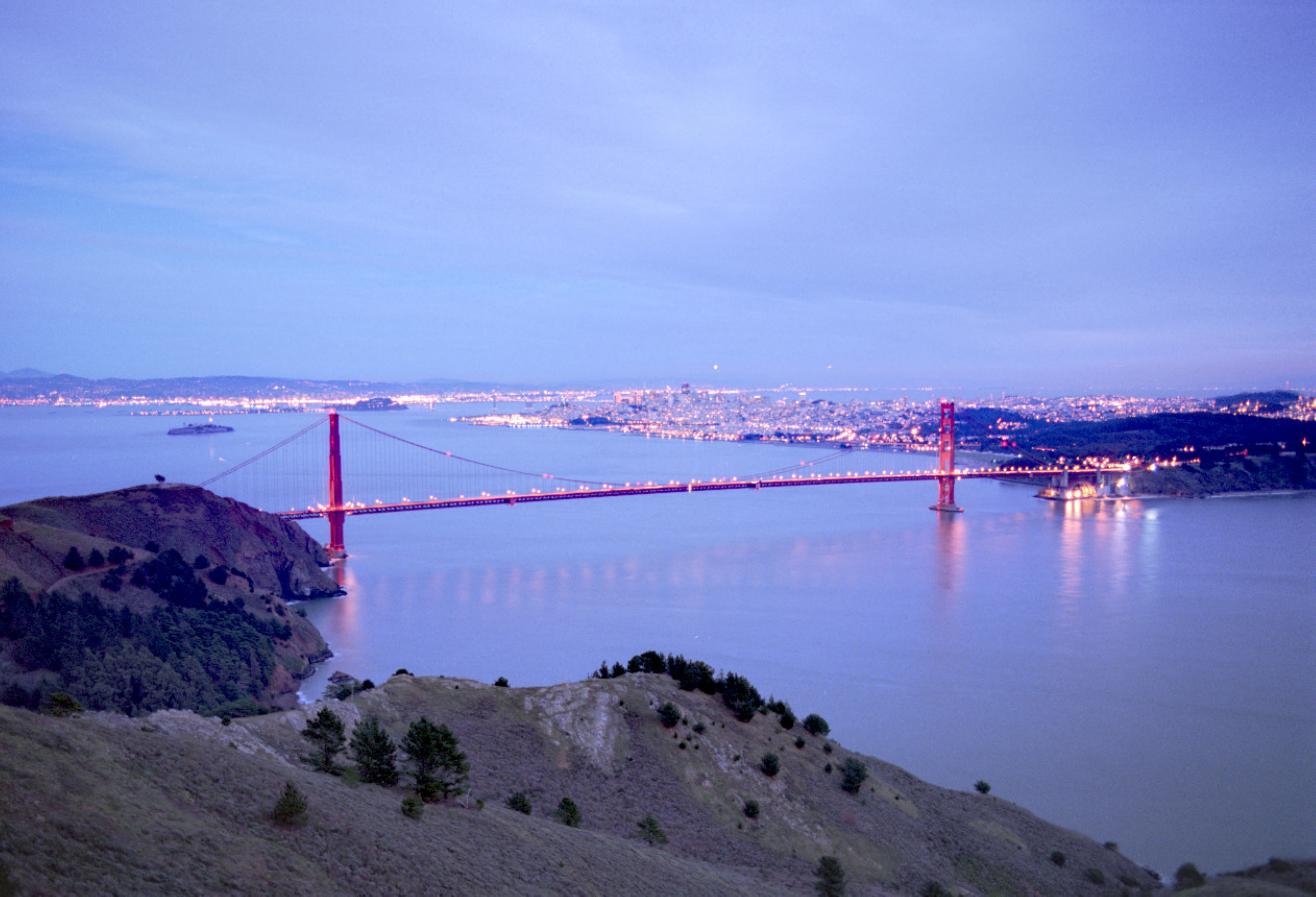}
    \label{fig:ProposedMertens}}
  \subfloat{
    \includegraphics[width=0.32\hsize]{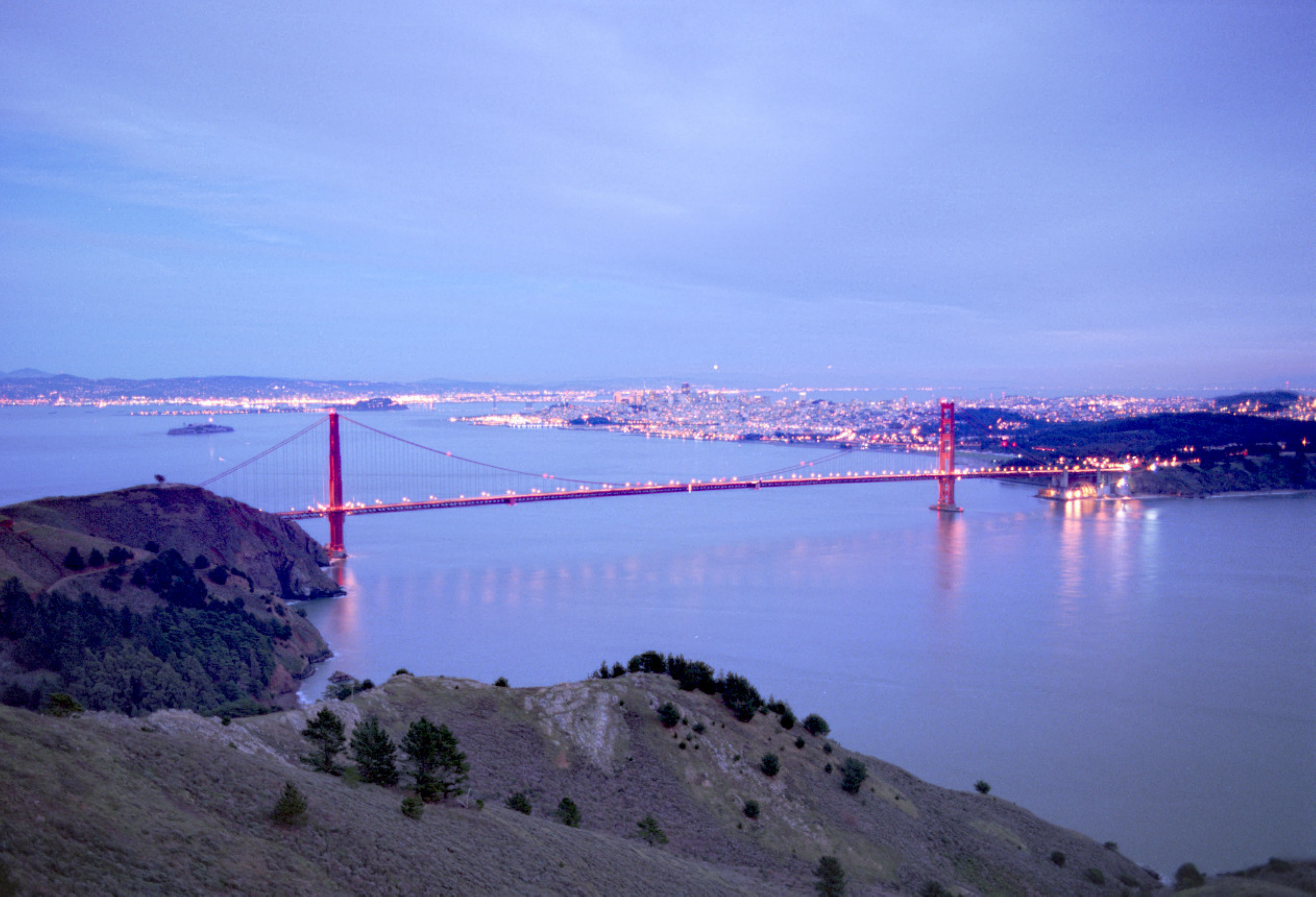}
    \label{fig:ProposedSakai}}
  \subfloat{
    \includegraphics[width=0.32\hsize]{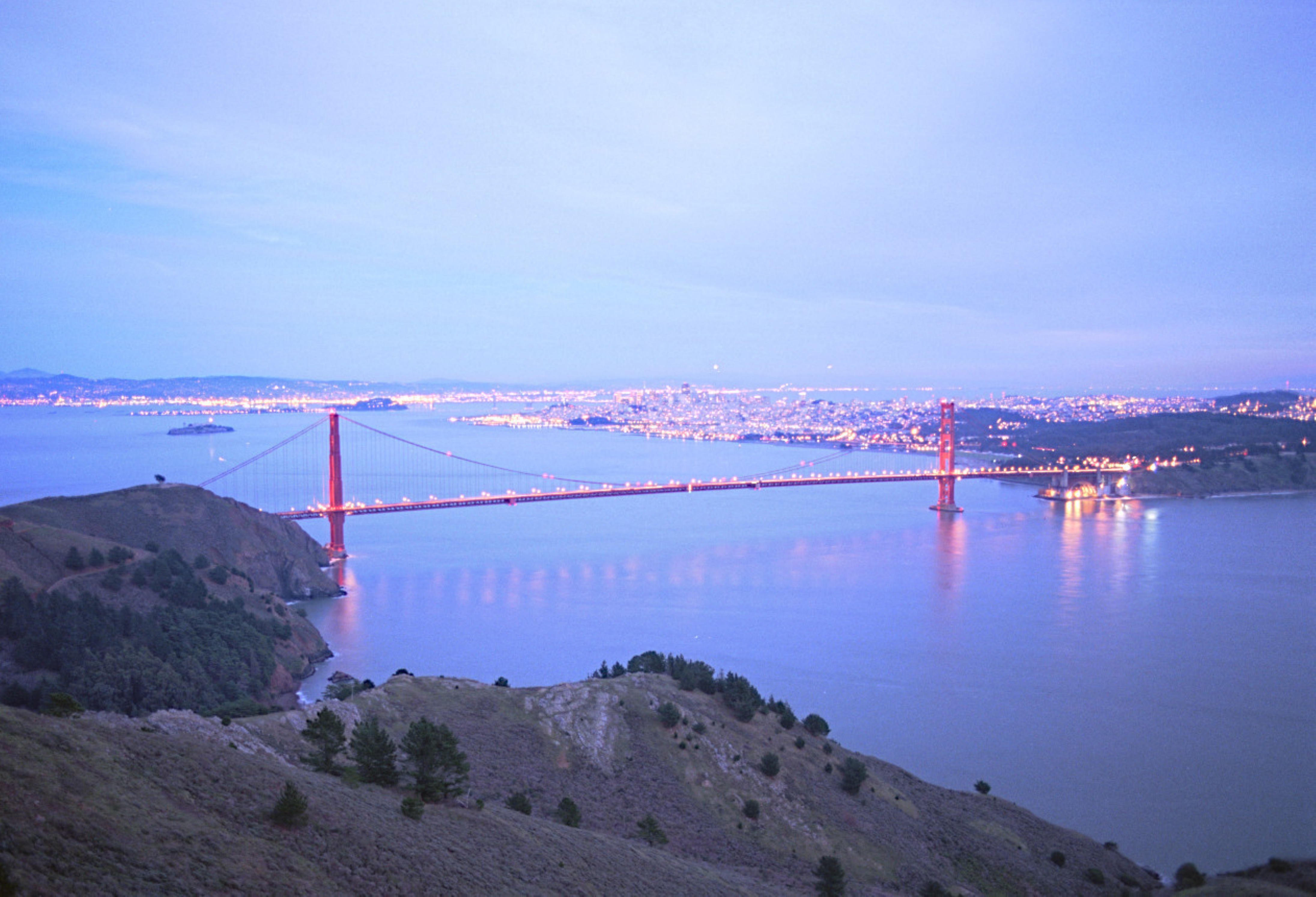}
    \label{fig:ProposedNejati}}\\
  \vspace{2pt}
  \addtocounter{subfigure}{-3}
  \subfloat[Proposed with \cite{mertens2009exposure}]{
    \includegraphics[width=0.32\hsize]{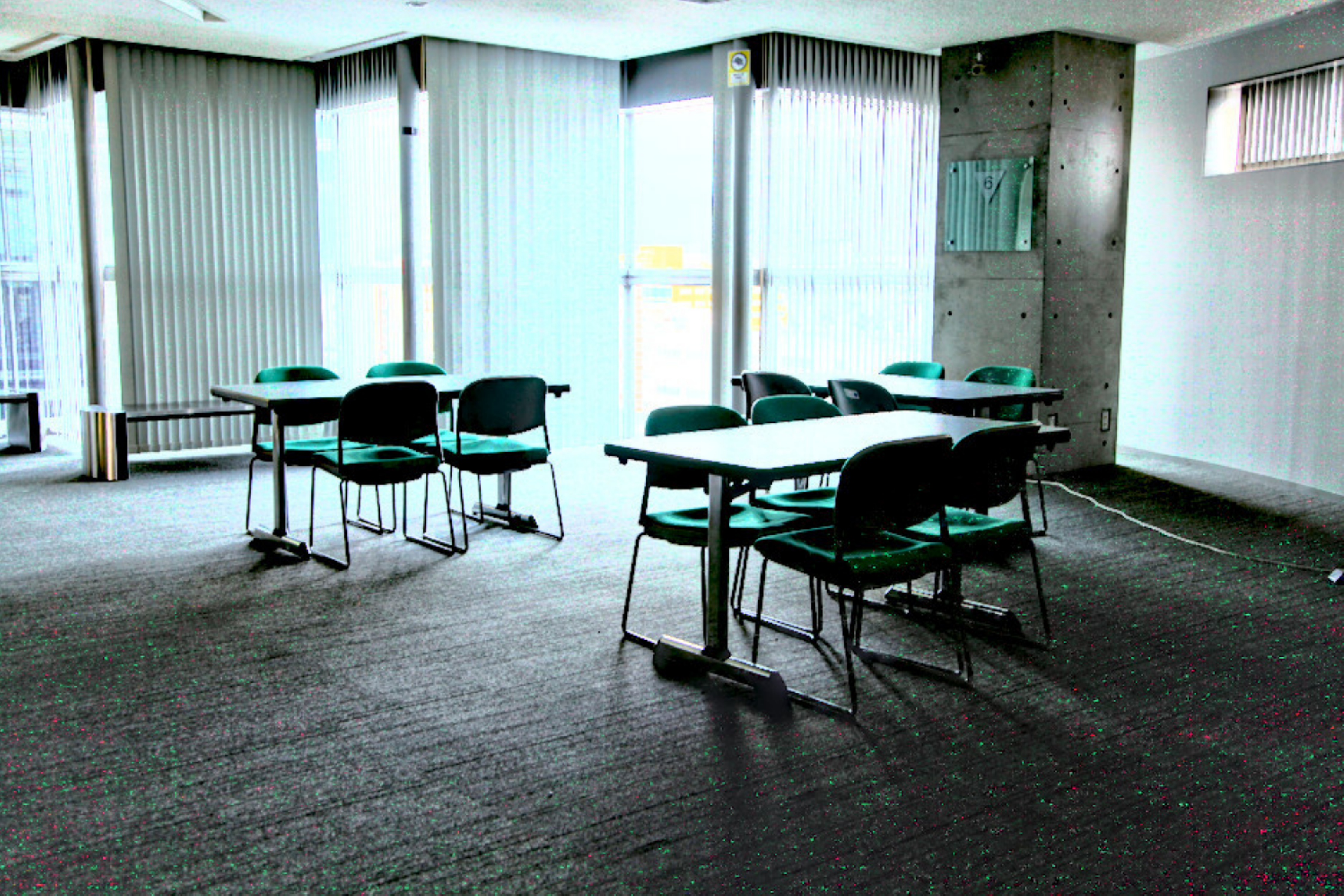}
    \label{fig:Lobby_PMEFMertens}}
  \subfloat[Proposed with \cite{sakai2015hybrid}]{
    \includegraphics[width=0.32\hsize]{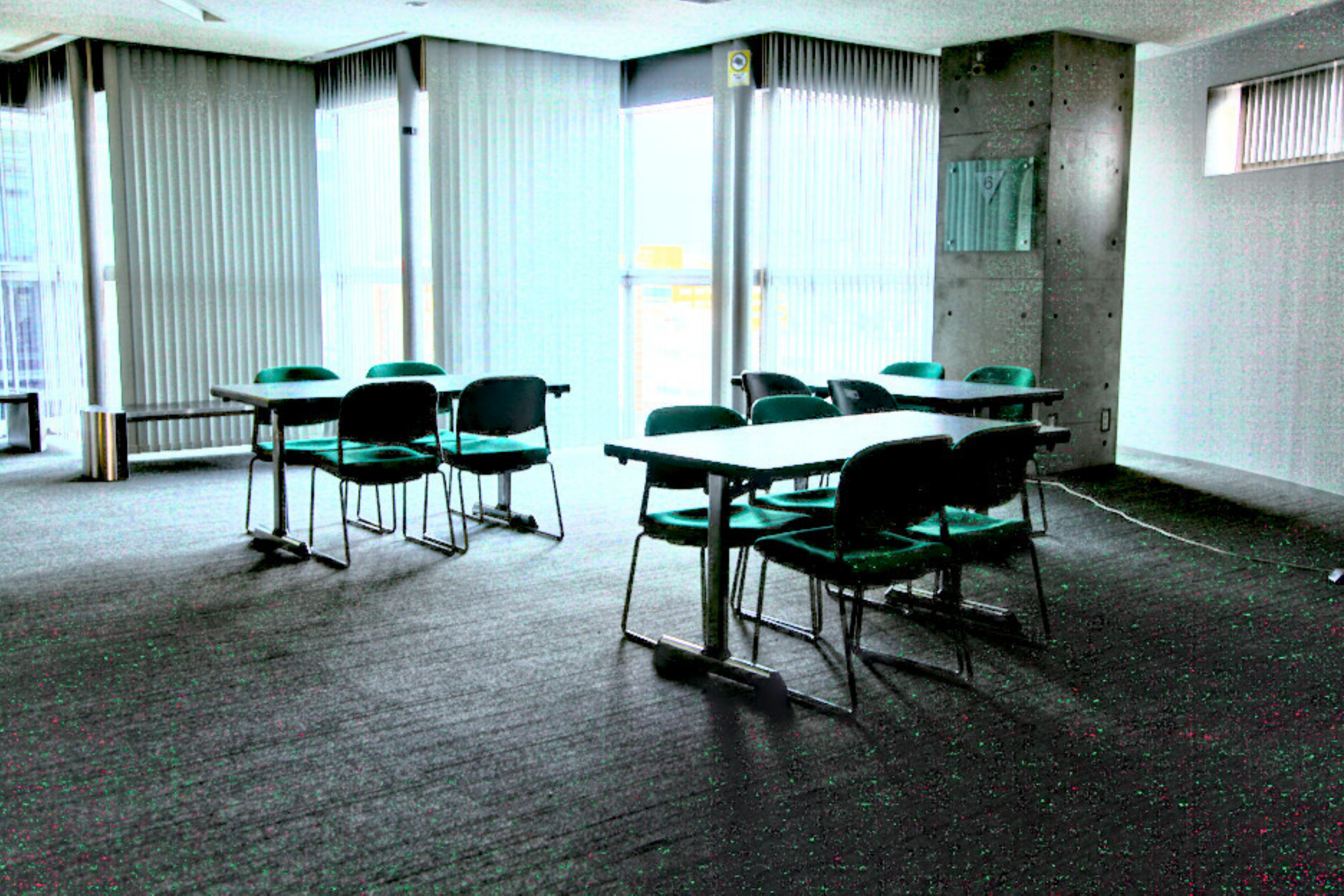}
    \label{fig:Lobby_PMEFSakai}}
  \subfloat[Proposed with \cite{nejati2017fast}]{
    \includegraphics[width=0.32\hsize]{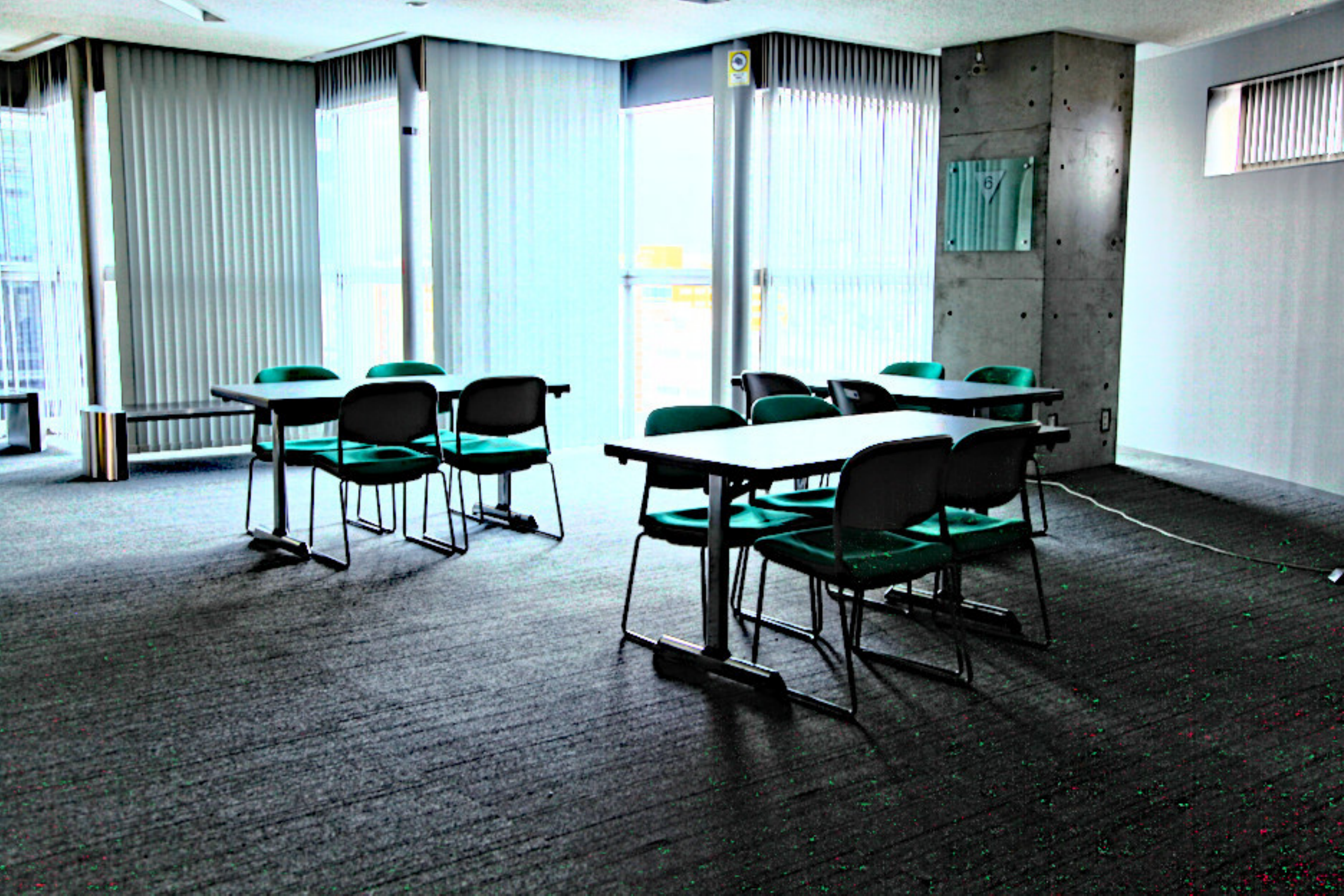}
    \label{fig:Lobby_PMEFNejati}}\\
  \caption{Images $I_f$ fused by six methods}
  \label{fig:resultsCamera}
\end{figure}

  Figure \ref{fig:resultsCamera} shows images $I_f$
  produced by each multi-exposure image fusion method.
  Tables \ref{tab:HDRTMQI}, \ref{tab:HDRNaturalness}, and \ref{tab:HDRDiscreteEntropy}
  summarize the TMQI score, statistical naturalness score, discrete entropy score
  for Simulation 1, respectively.
  For TMQI $\in [0, 1]$ (and statistical naturalness $\in [0, 1]$),
  a larger value means higher quality.
  The results in Table \ref{tab:HDRTMQI} indicate that
  the proposed method improves the quality of the produced images.
  Tables \ref{tab:HDRNaturalness} and \ref{tab:HDRDiscreteEntropy}
  show trends similar to that in Table \ref{tab:HDRTMQI}.
  
  These results demonstrate that the proposed method allows us to generate images with
  higher quality than that of one generated by original
  multi-exposure image fusion methods.
\subsubsection{Simulation 2}
  Tables \ref{tab:EOS5DNaturalness} and \ref{tab:EOS5DDiscreteEntropy}
  likewise show the results for Simulation 2.
  As shown in Table \ref{tab:EOS5DNaturalness},
  the statistical naturalness scores
  for conventional methods were extremely low
  when the original image was completely dark, like that shown in Fig.
  \ref{fig:inputImages}\subref{fig:LobbyP1EV}.
  The results mean that most areas in the resulting images are not viewable.
  
  Using the proposed method enhanced the brightness and contrast of the input
  images (see Fig. \ref{fig:inputImages}\subref{fig:EnhancedLobby}).
  This increased the statistical naturalness scores
  significantly (see Table \ref{tab:EOS5DNaturalness}).
  
  Table \ref{tab:EOS5DDiscreteEntropy}
  also shows a trend similar to that in Table \ref{tab:EOS5DNaturalness}.
  These results also demonstrate the proposed method enables us
  to produce images with higher quality,
  than conventional fusion methods.
\begin{table}[!t]
  \centering
  \caption{Experimental results for Simulation 1 (TMQI).
  Boldface indicates the higher score.}
  \scalebox{0.85}{\small
    \begin{tabular}{l|c|cc|cc|cc} \hline \hline
      Method & Input & \multicolumn{2}{|c|}{Mertens\cite{mertens2009exposure}} &
      \multicolumn{2}{|c|}{Sakai\cite{sakai2015hybrid}} &
      \multicolumn{2}{|c}{Nejati\cite{nejati2017fast}}\\
         & image & Original & Proposed & Original & Proposed & Original & Proposed\\ \hdashline
      Adjuster & 0.8793 & 0.8879 & \bfseries{0.9092} & 0.8907 & \bfseries{0.9122} &
      0.9053 & \bfseries{0.9214}\\
      ApartmentFloat & 0.7471 & 0.8331 & \bfseries{0.8462} & 0.8297 & \bfseries{0.8439} &
      0.8250 & \bfseries{0.8436}\\
      Balls & 0.8371 & 0.8269 & \bfseries{0.8300} & 0.8419 & \bfseries{0.8460} &
      0.8184 & \bfseries{0.8204}\\
      BrightRings & 0.7748 & 0.7771 & \bfseries{0.7868} & 0.7773 & \bfseries{0.7872} &
      0.7855 & \bfseries{0.8091}\\
      Cannon & 0.8412 & 0.9203 & \bfseries{0.9205} & 0.9336 & \bfseries{0.9340} &
      \bfseries{0.9039} & 0.9031\\
      GoldenGate & 0.7754 & 0.7911 & \bfseries{0.8030} & 0.7915 & \bfseries{0.8029} &
      0.8269 & \bfseries{0.8432}\\ \hdashline
      Average & \multirow{2}{*}{0.8086} & \multirow{2}{*}{0.8531} &
      \multirow{2}{*}{\bfseries{0.8695}} & \multirow{2}{*}{0.8536} &
      \multirow{2}{*}{\bfseries{0.8703}} & \multirow{2}{*}{0.8574} &
      \multirow{2}{*}{\bfseries{0.8713}}\\
      (60 images) & & & & & & &\\
      \hline
  \end{tabular}
  }
  \label{tab:HDRTMQI}
\end{table}
\begin{table}[!t]
  \centering
  \caption{Experimental results for Simulation 1 (statistical naturalness).
  Boldface indicates the higher score.}
  \scalebox{0.85}{\small
    \begin{tabular}{l|c|cc|cc|cc} \hline \hline
      Method & Input & \multicolumn{2}{|c|}{Mertens\cite{mertens2009exposure}} &
      \multicolumn{2}{|c|}{Sakai\cite{sakai2015hybrid}} &
      \multicolumn{2}{|c}{Nejati\cite{nejati2017fast}}\\
         & image & Original & Proposed & Original & Proposed & Original & Proposed\\ \hdashline
      Adjuster & 0.4350 & 0.4974 & \bfseries{0.5934} & 0.5129 & \bfseries{0.6106} &
      0.5494 & \bfseries{0.6279}\\
      ApartmentFloat & 0.2750 & \bfseries{0.4541} & 0.4240 & \bfseries{0.4337} & 0.4112 &
      \bfseries{0.4341} & 0.4269\\
      Balls & 0.1145 & 0.0930 & \bfseries{0.1046} & 0.1503 & \bfseries{0.1677} &
      0.0630 & \bfseries{0.0699}\\
      BrightRings & 0.0224 & 0.0340 & \bfseries{0.0780} & 0.0342 & \bfseries{0.0788} &
      0.0730 & \bfseries{0.1891}\\
      Cannon & 0.1051 & 0.4956 & \bfseries{0.4967} & 0.5760 & \bfseries{0.5786} &
      \bfseries{0.4028} & 0.3985\\
      GoldenGate & 0.0351 & 0.0755 & \bfseries{0.0879} & 0.0767 & \bfseries{0.0892} &
      0.1902 & \bfseries{0.2348}\\
      \hdashline
      Average & \multirow{2}{*}{0.2055} & \multirow{2}{*}{0.3717} &
      \multirow{2}{*}{\bfseries{0.4374}} & \multirow{2}{*}{0.3751} &
      \multirow{2}{*}{\bfseries{0.4421}} & \multirow{2}{*}{0.3919} &
      \multirow{2}{*}{\bfseries{0.4439}}\\
      (60 images) & & & & & & &\\ \hline
  \end{tabular}
  }
  \label{tab:HDRNaturalness}
\end{table}
\begin{table}[!t]
  \centering
  \caption{Experimental results for Simulation 1 (discrete entropy).
  Boldface indicates the higher score.}
  \scalebox{0.85}{\small
    \begin{tabular}{l|c|cc|cc|cc} \hline \hline
      Method & Input & \multicolumn{2}{|c|}{Mertens\cite{mertens2009exposure}} &
      \multicolumn{2}{|c|}{Sakai\cite{sakai2015hybrid}} &
      \multicolumn{2}{|c}{Nejati\cite{nejati2017fast}}\\
         & image & Original & Proposed & Original & Proposed & Original & Proposed\\ \hdashline
      Adjuster & 7.225 & 7.520 & \bfseries{7.616} & 7.531 & \bfseries{7.625} &
      7.646 & \bfseries{7.754}\\
      ApartmentFloat & 5.648 & 6.315 & \bfseries{7.211} & 6.328 & \bfseries{7.219} &
      6.191 & \bfseries{6.751}\\ 
      Balls & 7.412 & 7.723 & \bfseries{7.745} & 7.736 & \bfseries{7.754} &
      7.772 & \bfseries{7.794}\\
      BrightRings & 1.491 & 2.070 & \bfseries{2.280} & 2.262 & \bfseries{2.484} &
      \bfseries{2.807} & 2.707\\
      Cannon & 6.913 & 7.566 & \bfseries{7.568} & \bfseries{7.556} & \bfseries{7.556} &
      7.521 & \bfseries{7.525}\\
      GoldenGate & 7.235 & 7.127 & \bfseries{7.129} & 7.136 & \bfseries{7.137} &
      7.477 & \bfseries{7.501}\\ \hdashline
      Average & \multirow{2}{*}{6.017} & \multirow{2}{*}{6.572} &
      \multirow{2}{*}{\bfseries{6.730}} & \multirow{2}{*}{6.656} &
      \multirow{2}{*}{\bfseries{6.787}} & \multirow{2}{*}{6.633} &
      \multirow{2}{*}{\bfseries{6.755}}\\
      (60 images) & & & & & & &\\ \hline
  \end{tabular}
  }
  \label{tab:HDRDiscreteEntropy}
\end{table}
\begin{table}[!t]
  \centering
  \caption{Experimental results for Simulation 2 (statistical naturalness).
  Boldface indicates the higher score.}
  \scalebox{0.85}{\small
    \begin{tabular}{l|c|cc|cc|cc} \hline \hline
      Method & Input & \multicolumn{2}{|c|}{Mertens\cite{mertens2009exposure}} &
      \multicolumn{2}{|c|}{Sakai\cite{sakai2015hybrid}} &
      \multicolumn{2}{|c}{Nejati\cite{nejati2017fast}}\\
         & image & Original & Proposed & Original & Proposed & Original & Proposed\\ \hdashline
      Corridor & 0.0000 & 0.0000 & \bfseries{0.3750} & 0.0000 & \bfseries{0.3736} &
      0.0000 & \bfseries{0.3637}\\
      Lobby & 0.0006 & 0.0063 & \bfseries{0.5988} & 0.0058 & \bfseries{0.6035} &
      0.0054 & \bfseries{0.5541}\\
      Window & 0.0044 & 0.0333 & \bfseries{0.4679} & 0.0308 & \bfseries{0.4644} &
      0.0261 & \bfseries{0.5080}\\ \hline
  \end{tabular}
  }
  \label{tab:EOS5DNaturalness}
\end{table}
\begin{table}[!t]
  \centering
  \caption{Experimental results for Simulation 2 (discrete entropy).
  Boldface indicates the higher score.}
  \scalebox{0.85}{\small
    \begin{tabular}{l|c|cc|cc|cc} \hline \hline
      Method & Input & \multicolumn{2}{|c|}{Mertens\cite{mertens2009exposure}} &
      \multicolumn{2}{|c|}{Sakai\cite{sakai2015hybrid}} &
      \multicolumn{2}{|c}{Nejati\cite{nejati2017fast}}\\
         & image & Original & Proposed & Original & Proposed & Original & Proposed\\ \hdashline
      Corridor & 1.956 & 3.294 & \bfseries{7.524} & 3.316 & \bfseries{7.535} &
      2.654 & \bfseries{7.533}\\
      Lobby & 3.016 & 5.274 & \bfseries{7.372} & 5.301 & \bfseries{7.392} &
      5.035 & \bfseries{7.257}\\
      Window & 5.173 & 5.851 & \bfseries{7.265} & 5.868 & \bfseries{7.271} &
      5.822 & \bfseries{7.196}\\ \hline
  \end{tabular}
  }
  \label{tab:EOS5DDiscreteEntropy}
\end{table}
\section{Conclusion}
  This paper has proposed a novel multi-exposure image fusion method
  based on exposure compensation.
  The proposed method enhances the quality of input multi-exposure images
  by using local contrast enhancement, exposure compensation and tone mapping.
  To improve their quality, the proposed method
  utilizes the relationship between exposure values and pixel values.
  Experimental results showed that the proposed method can produce images with higher quality,
  than conventional multi-exposure image fusion methods,
  in terms of TMQI, statistical naturalness, and discrete entropy.


\begin{thebibliography}{10}
\providecommand{\url}[1]{#1}
\csname url@samestyle\endcsname
\providecommand{\newblock}{\relax}
\providecommand{\bibinfo}[2]{#2}
\providecommand{\BIBentrySTDinterwordspacing}{\spaceskip=0pt\relax}
\providecommand{\BIBentryALTinterwordstretchfactor}{4}
\providecommand{\BIBentryALTinterwordspacing}{\spaceskip=\fontdimen2\font plus
\BIBentryALTinterwordstretchfactor\fontdimen3\font minus
  \fontdimen4\font\relax}
\providecommand{\BIBforeignlanguage}[2]{{%
\expandafter\ifx\csname l@#1\endcsname\relax
\typeout{** WARNING: IEEEtran.bst: No hyphenation pattern has been}%
\typeout{** loaded for the language `#1'. Using the pattern for}%
\typeout{** the default language instead.}%
\else
\language=\csname l@#1\endcsname
\fi
#2}}
\providecommand{\BIBdecl}{\relax}
\BIBdecl

\bibitem{zuiderveld1994contrast}
K.~Zuiderveld, ``Contrast limited adaptive histograph equalization,'' in
  \emph{Graphics gems IV}.\hskip 1em plus 0.5em minus 0.4em\relax Academic
  Press Professional, Inc., 1994, pp. 474--485.

\bibitem{wu2017contrast}
X.~Wu, X.~Liu, K.~Hiramatsu, and K.~Kashino, ``Contrast-accumulated histogram
  equalization for image enhnacement,'' in \emph{2017 International Conference
  on Image Processing (ICIP)}.\hskip 1em plus 0.5em minus 0.4em\relax IEEE,
  2017, pp. 3190--3194.

\bibitem{kinoshita2017pseudo}
Y.~Kinoshita, T.~Yoshida, S.~Shiota, and H.~Kiya, ``Pseudo multi-exposure
  fusion using a single image,'' in \emph{APSIPA Annual Summit and Conference},
  2017, pp. 263--269.

\bibitem{goshtasby2005fusion}
A.~A. Goshtasby, ``Fusion of multi-exposure images,'' \emph{Image and Vision
  Computing}, vol.~23, no.~6, pp. 611--618, 2005.

\bibitem{mertens2009exposure}
T.~Mertens, J.~Kautz, and F.~Van~Reeth, ``Exposure fusion: A simple and
  practical alternative to high dynamic range photography,'' \emph{Computer
  Graphics Forum}, vol.~28, no.~1, pp. 161--171, 2009.

\bibitem{saleem2012image}
A.~Saleem, A.~Beghdadi, and B.~Boashash, ``Image fusion-based contrast
  enhancement,'' \emph{EURASIP Journal on Image and Video Processing}, vol.
  2012, no.~1, p.~10, 2012.

\bibitem{wang2015exposure}
J.~Wang, G.~Xu, and H.~Lou, ``Exposure fusion based on sparse coding in pyramid
  transform domain,'' in \emph{Proceedings of the 7th International Conference
  on Internet Multimedia Computing and Service}, ser. ICIMCS '15.\hskip 1em
  plus 0.5em minus 0.4em\relax New York, NY, USA: ACM, 2015, pp. 4:1--4:4.

\bibitem{li2014selectively}
Z.~Li, J.~Zheng, Z.~Zhu, and S.~Wu, ``Selectively detail-enhanced fusion of
  differently exposed images with moving objects,'' \emph{IEEE Transactions on
  Image Processing}, vol.~23, no.~10, pp. 4372--4382, 2014.

\bibitem{sakai2015hybrid}
T.~Sakai, D.~Kimura, T.~Yoshida, and M.~Iwahashi, ``Hybrid method for
  multi-exposure image fusion based on weighted mean and sparse
  representation,'' in \emph{2015 23rd European Signal Processing Conference
  (EUSIPCO)}.\hskip 1em plus 0.5em minus 0.4em\relax EURASIP, 2015, pp.
  809--813.

\bibitem{nejati2017fast}
M.~Nejati, M.~Karimi, S.~M.~R. Soroushmehr, N.~Karimi, S.~Samavi, and
  K.~Najarian, ``Fast exposure fusion using exposedness function,'' in
  \emph{2017 International Conference on Image Processing (ICIP)}.\hskip 1em
  plus 0.5em minus 0.4em\relax IEEE, 2017, pp. 2234--2238.

\bibitem{debevec1997recovering}
P.~E. Debevec and J.~Malik, ``Recovering high dynamic range radiance maps from
  photographs,'' in \emph{ACM SIGGRAPH}.\hskip 1em plus 0.5em minus 0.4em\relax
  ACM, 1997, pp. 369--378.

\bibitem{reinhard2002photographic}
E.~Reinhard, M.~Stark, P.~Shirley, and J.~Ferwerda, ``Photographic tone
  reproduction for digital images,'' \emph{ACM Transactions on Graphics (TOG)},
  vol.~21, no.~3, pp. 267--276, 2002.

\bibitem{oh2015robust}
T.-H. Oh, J.-Y. Lee, Y.-W. Tai, and I.~S. Kweon, ``Robust high dynamic range
  imaging by rank minimization,'' \emph{IEEE Transactions on Pattern Analysis
  and Machine Intelligence}, vol.~37, no.~6, pp. 1219--1232, 2015.

\bibitem{kinoshita2016remapping}
Y.~Kinoshita, S.~Shiota, M.~Iwahashi, and H.~Kiya, ``An remapping operation
  without tone mapping parameters for hdr images,'' \emph{IEICE Transactions on
  Fundamentals of Electronics, Communications and Computer Sciences}, vol.~99,
  no.~11, pp. 1955--1961, 2016.

\bibitem{kinoshita2017fast}
Y.~Kinoshita, S.~Shiota, and H.~Kiya, ``Fast inverse tone mapping with
  reinhard's grobal operator,'' in \emph{2017 IEEE International Conference on
  Acoustics, Speech and Signal Processing (ICASSP)}.\hskip 1em plus 0.5em minus
  0.4em\relax IEEE, 2017, pp. 1972--1976.

\bibitem{kinoshita2017fast_trans}
------, ``Fast inverse tone mapping based on reinhard's global operator with
  estimated parameters,'' \emph{IEICE Transactions on Fundamentals of
  Electronics, Communications and Computer Sciences}, vol. 100, no.~11, pp.
  2248--2255, 2017.

\bibitem{huo2016single}
Y.~Q. Huo and X.~D. Zhang, ``Single image-based hdr imaging with crf
  estimation,'' in \emph{2016 International Conference On Communication
  Problem-Solving (ICCP)}.\hskip 1em plus 0.5em minus 0.4em\relax IEEE, 2016,
  pp. 1--3.

\bibitem{murofushi2013integer}
T.~Murofushi, M.~Iwahashi, and H.~Kiya, ``An integer tone mapping operation for
  hdr images expressed in floating point data,'' in \emph{2013 IEEE
  International Conference on Acoustics, Speech and Signal Processing
  (ICASSP)}.\hskip 1em plus 0.5em minus 0.4em\relax IEEE, 2013, pp. 2479--2483.

\bibitem{murofushi2014integer}
T.~Murofushi, T.~Dobashi, M.~Iwahashi, and H.~Kiya, ``An integer tone mapping
  operation for hdr images in openexr with denormalized numbers,'' in
  \emph{2014 IEEE International Conference on Image Processing (ICIP)}.\hskip
  1em plus 0.5em minus 0.4em\relax IEEE, 2014, pp. 4497--4501.

\bibitem{dobashi2014fixed}
T.~Dobashi, T.~Murofushi, M.~Iwahashi, and K.~Hitoshi, ``A fixed-point global
  tone mapping operation for hdr images in the rgbe format,'' \emph{IEICE
  Transactions on Fundamentals of Electronics, Communications and Computer
  Sciences}, vol.~97, no.~11, pp. 2147--2153, 2014.

\bibitem{dufaux2016high}
F.~Dufaux, P.~L. Callet, R.~Mantiuk, and M.~Mrak, \emph{High Dynamic Range
  Video, From Acquisition, to Display and Applications}.\hskip 1em plus 0.5em
  minus 0.4em\relax Elsevier Ltd., 2016.

\bibitem{huo2013dodging}
H.~Youngquing, Y.~Fan, and V.~Brost, ``Dodging and burning inspired inverse
  tone mapping algorithm,'' \emph{Journal of Computational Information
  Systems}, vol.~9, no.~9, pp. 3461--3468, 2013.

\bibitem{yeganeh2013objective}
H.~Yeganeh and Z.~Wang, ``Objective quality assessment of tone mapped images,''
  \emph{IEEE Transactions on Image Processing}, vol.~22, no.~2, pp. 657--667,
  2013.

\bibitem{openexrimage}
\BIBentryALTinterwordspacing
``Github - openexr.'' [Online]. Available: \url{https://github.com/openexr/}
\BIBentrySTDinterwordspacing

\bibitem{anyherehdrimage}
\BIBentryALTinterwordspacing
``High dynamic range image examples.'' [Online]. Available:
  \url{\\http://www.anyhere.com/gward/hdrenc/pages/\\originals.html}
\BIBentrySTDinterwordspacing

\end{thebibliography}
\end{document}